\documentclass{article}
\usepackage{iclr2027_conference,times}

\usepackage{amsmath,amsfonts,bm}

\def\eqref#1{equation~\ref{#1}}

\def\1{\bm{1}}

\DeclareMathAlphabet{\mathsfit}{\encodingdefault}{\sfdefault}{m}{sl}
\SetMathAlphabet{\mathsfit}{bold}{\encodingdefault}{\sfdefault}{bx}{n}

\usepackage{url}
\usepackage{graphicx}
\usepackage[table]{xcolor}

\usepackage{booktabs}
\usepackage{multirow}
\usepackage{tabularx}
\usepackage{capt-of}
\usepackage{placeins}
\usepackage[most]{tcolorbox}
\usepackage[
  colorlinks=true,
  linkcolor=blue,
  citecolor=blue,
  urlcolor=blue
]{hyperref}

\newtcolorbox{promptbox}[2][]{
    enhanced,
    colback=yellow!8,
    colframe=yellow!45!black,
    colbacktitle=yellow!20,
    coltitle=black,
    fonttitle=\bfseries,
    boxrule=0.5pt,
    arc=2mm,
    left=8pt,
    right=8pt,
    top=7pt,
    bottom=7pt,
    before skip=8pt,
    after skip=8pt,
    title={#2},
    #1
}

\newcommand{\speechcritic}{\textsc{SpeechCritic}}

\title{SpeechCritic: Learning a Diagnostic Speech \\ Judge from Limited Human Preferences}

\author{
\makebox[\dimexpr\textwidth-2\tabcolsep\relax][c]{%
Mingyue Huo$^{1}$\thanks{Work done during internship at Netflix.}\quad
Shivam Mehta$^{2}$\quad Bhavin Jawade$^{2}$\quad Yinghong Lan$^{2}$\quad Haoqi Li$^{2}$}\\
\normalfont\makebox[\dimexpr\textwidth-2\tabcolsep\relax][c]{%
$^{1}$University of Illinois Urbana-Champaign \qquad $^{2}$Netflix}\\[-1pt]
\normalfont\small\makebox[\dimexpr\textwidth-2\tabcolsep\relax][c]{%
\texttt{mhuo5@illinois.edu} \qquad \texttt{haoqil@netflix.com}}
}

\iclrfinalcopy
\begin{document}

\vspace*{-0.25in}
\maketitle
\lhead{}
\renewcommand{\headrulewidth}{0pt}
\vspace{-24pt}

\begin{abstract}
Human speech conveys rich perceptual information, such as emotion and speaker identity, yet most automatic speech quality judges reduce it to a single naturalness score. We study \textit{diagnostic} speech judges: given two candidates, a diagnostic judge decides which is better, along which perceptual dimensions (e.g., timbre, emotion, timing) they differ, and which audible cues support its decision. Learning such judges is challenging: expert annotation is costly, and simply prompting a frontier audio-language model to produce labels is unreliable: our probing reveals substantial errors and unstable instruction following.
We introduce \speechcritic{}, which learns a diagnostic judge in a reference-conditioned cross-lingual setting from only about 300 human-labeled comparisons. Rather than replacing the frontier model, \speechcritic{} calibrates it with these labels: for each dimension, it selects the acoustic measurements that agree with human judgments, maps them to A/\textsc{Tie}/B probabilities, and passes these to the model as non-binding hints alongside the audio. Compared with the same model labeling without hints, this raises dimension-level agreement with humans by 6.3 points and cuts the mismatch with human \textsc{Tie} rates by 10.4 points.
We then train a 7B judge on this supervision and find that different training signals shape different judge behaviors: supervised fine-tuning establishes the task, on-policy distillation transfers the teacher's dimension-level strengths and weaknesses, and reinforcement learning helps most on clear-cut comparisons where human raters agree.
Notably, human listeners also find that reinforcement learning makes rationales cite more specific, localized acoustic cues, although it never directly rewards rationale text. 
Finally, we show that the pipeline is language-pair agnostic by instantiating it on both English–Japanese and English–Spanish.
Together, these results demonstrate a path from limited human preferences to a diagnostic speech judge. Code and Demo are available at \url{https://github.com/mingyue66/Speech_Critic}.
\end{abstract}
\vspace{-10pt}

\begin{figure}[h!]
    \centering
    \includegraphics[
        width=\textwidth,
        trim=84bp 184bp 99bp 170bp,
        clip
    ]{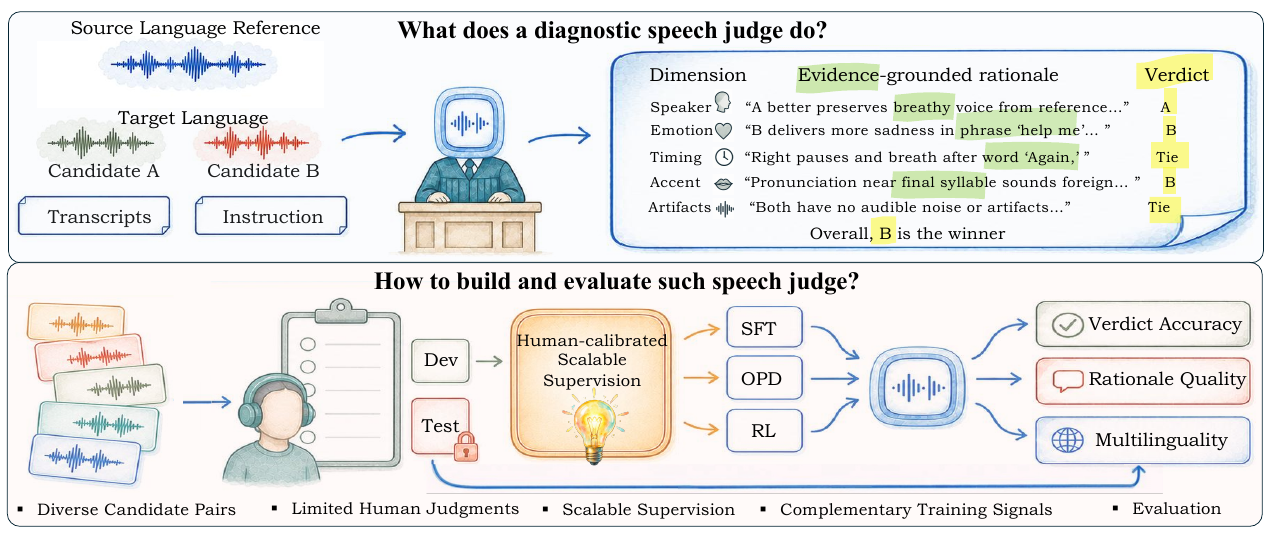}
    \caption{In a reference-conditioned cross-lingual comparison,
    a diagnostic speech judge produces multidimensional verdicts and evidence-grounded rationales that identify audible differences between two candidates.
    Our \speechcritic{} framework covers the end-to-end data, training, and evaluation pipeline. The central
    human-calibration method is introduced in Section~\ref{sec:method}.}
    \label{fig:overview}
\end{figure}

\section{Introduction}
\label{sec:introduction}

Recent multimodal large language models can directly process speech, making automatic speech evaluation increasingly plausible. Yet speech perception is inherently multidimensional, and a useful evaluator should explain its preference with evidence rather than merely select a winner. We study diagnostic speech judging for pairwise speech comparison. A diagnostic judge listens to two speech candidates, whether natural or synthesized, and produces an Overall verdict, judgments across multiple perceptual dimensions, and evidence-grounded rationales supported by audibly verifiable cues. We instantiate this general task in a reference-conditioned cross-lingual setting, where the judge compares two target-language candidates against a source-language reference in terms of speaker characteristics, emotion, timing, pronunciation or accent, and audio artifacts.

Building such a judge is challenging because human listeners may reasonably disagree on subtle perceptual differences, while collecting consistent dimension-level judgments and audio-grounded explanations requires substantial expert effort. Existing approaches either train speech judges on large human preference corpora that are costly and time-consuming to collect~\citep{speechjudge} or directly prompt frontier audio-language models as evaluators~\citep{audiojudge}. To examine the latter option, we probed three proprietary and five open-source models. Their results reveal weak human agreement, skewed candidate or \textsc{Tie} preferences, and inconsistent dimension-level judgments, making their outputs unreliable as training supervision. This raises our central question:

\begin{quote}
\emph{Can limited human judgments be used to build a scalable, specialized diagnostic speech judge without collecting human labels at training scale?}
\end{quote}

Our key idea is to use human judgments to calibrate machine supervision rather than replace it. \speechcritic{} follows a select--calibrate--scale pipeline. Select: limited human judgments determine which automatic acoustic measurements provide reliable evidence for each supported dimension. Calibrate: each retained signal is mapped to uncertainty-aware probabilities over \(A\), \textsc{Tie}, and \(B\). Scale: the frozen mappings provide non-binding hints to a machine labeler, which still listens to the original speech and generates verdict-and-rationale supervision for more than 10,000 comparisons. Unlike conventional weak supervision, these measurements neither determine the labels nor replace the raw audio~\citep{snorkel}. Unlike approaches based on human-written principles~\citep{constitutional_ai} or recursively generated model judgments~\citep{self_taught_evaluator}, humans directly determine which acoustic evidence the machine labeler should trust. This calibration improves agreement with human dimension-level judgments by 6.3 percentage points and reduces the mismatch in \textsc{Tie} behavior by 10.4 points.

Building scalable supervision is only part of the problem. We further study how that supervision should be used by comparing supervised fine-tuning (SFT), on-policy distillation (OPD), and reinforcement learning (RL). SFT establishes the diagnostic task, but its gains are not monotonic in the amount of supervision, and adapting the audio path improves overall decisions without uniformly strengthening dimensional diagnosis. OPD transfers the teacher’s judgment profile, with teacher behavior and conditioning mattering more than student initialization.  RL sharpens the verdict behavior encoded by its reward, with gains concentrated on high-consensus comparisons and a trade-off in dimensional diagnosis. Human evaluation further shows that RL can improve acoustic grounding without direct rationale rewards. Separately, persuasive rationales can still accompany incorrect verdicts. Finally, both the \speechcritic{} supervision pipeline and the resulting judges transfer across Japanese and Spanish, demonstrating that the framework extends beyond a single target language.

Figure~\ref{fig:overview} summarizes our \speechcritic{} framework. Our contributions are:

\begin{itemize}
    \item \textbf{Diagnostic task and human evaluation.} We formulate diagnostic speech judging with a five-dimensional rubric, analyze human subjectivity and agreement, and systematically probe zero-shot audio-language models.

    \item \textbf{Human-calibrated scalable supervision.} We introduce a select--calibrate--scale pipeline that turns roughly 300 human-labeled comparisons into more than 10,000 verdict-and-rationale training comparisons.

    \item \textbf{Learning and evaluation.} We characterize how SFT, OPD, and RL learn differently from imperfect supervision, evaluate the evidence grounding and reliability of their rationales with human listeners, and study how the framework transfers across target languages.
\end{itemize}

\section{Related Work}
\label{sec:related}

\paragraph{Audio-language model judges.}
Recent work follows two main routes: prompting existing audio-language models and training dedicated evaluators.
AudioJudge directly prompts off-the-shelf audio models~\citep{audiojudge}, while TRACE converts acoustic cues into textual descriptions for reasoning by a text LLM~\citep{trace}.
Dedicated evaluators and supporting resources cover instruction-driven assessment, low-level speech quality, multilingual judging, spoken dialogue, and multi-task evaluation~\citep{jastin,qualispeech,sqllm,wavreward,unisrm}.
SpeechJudge and GSRM further train specialized judges from large-scale human preferences or expert ratings~\citep{speechjudge,gsrm}.
Much of this literature emphasizes naturalness or low-level acoustic quality. 
Complementary auditing work exposes another limitation: compared with human listeners, speech LLM judges can rely excessively on acoustic shortcuts such as intensity, content richness, and emotional delivery, while their rationales rarely reveal these influences~\citep{acoustic_shortcuts}.
Our task instead learns a diagnostic speech judge from limited human judgments, aiming to align its multidimensional decisions with human preferences and ground its rationales in audible evidence.

\vspace{-5pt}

\paragraph{Scalable evaluator training.}
The broader LLM literature has explored several ways to reduce large-scale human preference annotation.
Prometheus and JudgeLM distill model-generated feedback into trainable text evaluators~\citep{prometheus,judgelm}, while Auto-J and Critique-out-Loud generate explicit critiques before final judgments or rewards~\citep{autoj,cloud}.
RLAIF replaces human preferences with AI feedback~\citep{rlaif}; Constitutional AI guides that feedback with human-written principles~\citep{constitutional_ai}; and Self-Taught Evaluators bootstrap evaluator reasoning from synthetic data~\citep{self_taught_evaluator}.
These studies show that evaluator supervision can be generated or distilled.
In speech modality, however, the supervising model must perceive the acoustic signal and ground both its verdicts and rationales in audible evidence, making the reliability of machine-generated supervision itself a central concern.

\paragraph{Weak supervision and domain evidence.}
Speech evaluation has long relied on specialized measurements, including speaker-embedding similarity~\citep{wespeaker}, emotion representations~\citep{wagner2023emotion,emotion2vec}, temporal and pronunciation measures, and neural quality predictors~\citep{utmos,dnsmos,nisqa}.
These tools estimate individual or narrowly defined perceptual properties, but do not jointly produce comparative verdicts and audio-grounded rationales.
Our use of such measurements is related to programmatic weak supervision, where systems such as Snorkel combine noisy heuristic labeling functions into probabilistic training labels~\citep{snorkel}.
Our measurements, however, do not directly label comparisons or replace the raw speech.
Although TRACE and GSRM also expose structured acoustic evidence to evaluators, we screen each candidate measurement against held-out human judgments for the corresponding dimension, calibrate retained signals into probabilistic $A$/\textsc{Tie}/$B$ hints, and provide them only as non-binding guidance to a raw-audio machine labeler.

\vspace{-5pt}

\section{Diagnostic Speech Judging: Task and Human Evaluation}
\label{sec:task}

\paragraph{Task formulation.}
\label{sec:sec:task_formulation}
We study \textit{reference-conditioned cross-lingual comparison}. 
Each comparison contains a source-language reference utterance $x^{\mathrm{src}}$
and two target-language speech candidates $x^{A}$ and $x^{B}$. Our primary experiments use an English reference and
Japanese candidates.
The judge also receives source and target transcripts with the matched semantic meaning ($t^{\mathrm{src}}$, $t^{\mathrm{tgt}}$) and an instruction specifying the rubric $r$.
We denote the full input by
\[
x = \left(x^{\mathrm{src}}, x^{A}, x^{B},
t^{\mathrm{src}}, t^{\mathrm{tgt}}, r\right),
\]

The judge outputs a structured natural language response following the instruction, which contains a verdict $v_k$ for each of five perceptual dimensions, and a binary Overall verdict $v_{\mathrm{overall}}$.
The output also contains rationales that support any verdict it makes, identifying localized and audibly verifiable observations, such as a word, phrase, pause, or change in prosodic delivery.
Such rationales can help move automated evaluation closer to the actionable feedback of an expert human listener.
\[
v_k \in \{A, B, \mathrm{Tie}\},\quad k=1,\ldots,5
\qquad\qquad
v_{\mathrm{overall}} \in \{A, B\}. 
\]

The five dimensions are Speaker, Emotion, Timing,
Pronunciation/Accent, and Audio Artifacts.
This rubric is empirically sufficient for the task: a simple equal-weight aggregation of human dimension-level votes recovers human Overall preferences with 93.0\% out-of-fold accuracy after a one-feature logistic calibration on the development set.

\paragraph{Constructing diverse data.}
\label{sec:sec:pair_construction}

We constructed a heterogeneous speech pool using multiple off-the-shelf speech synthesis systems. Their distinct behaviors and failure modes create natural variation across perceptual dimensions. For example, cross-lingual synthesis produces varying degrees of accent leakage, yielding candidate pairs that differ in Pronunciation/Accent quality. Appendix~\ref{app:data_construction} describes the pool in detail. We constructed more than 10,000 comparisons in total.

\paragraph{Human preference benchmark.}
\label{sec:sec:human_benchmark}
We recruited 20 native Japanese listeners who also understand English. Human listeners are given input $x$ and asked to produce verdicts $v_k$ and $v_\text{overall}$, without the need to write down any explanations. Each comparison receives three human judgments. In subsequent analyses, comparisons with unanimous $v_\text{overall}$ votes ($3$--$0$) are marked as
\emph{high-consensus comparisons} and those with split votes ($2$--$1$) as \emph{low-consensus comparisons}.

Human annotation yields 315 development comparisons and 290 test comparisons. Only the development set is used to calibrate the following supervision pipeline; the test set is never used for
calibration, training, or model selection. 
Throughout the paper, we evaluate the binary Overall verdict using accuracy over
$\{A,B\}$ and the five dimension-level verdicts using macro-F1 over
$\{A,B,\mathrm{Tie}\}$, averaged across dimensions. To contextualize the specific task's subjectivity, we  additionally collect all 20 listeners' preference for 30 anchor comparisons. Appendix~\ref{app:human_agreement} provides detailed annotation instructions, dimension definitions, and human agreement analysis.

On the full English--Japanese test set, the expected agreement of a randomly sampled panel listener with the panel-majority label is 85.8\% for $v_\text{overall}$ and 78.8\% for averaged dimensional macro-F1. We treat these values as label-reliability references rather than model-performance upper bounds.
\vspace{-10pt}

\paragraph{Zero-shot probing.}
\label{sec:sec:zero_shot}

\begin{figure}[h]
    \centering
    \includegraphics[width=\textwidth]{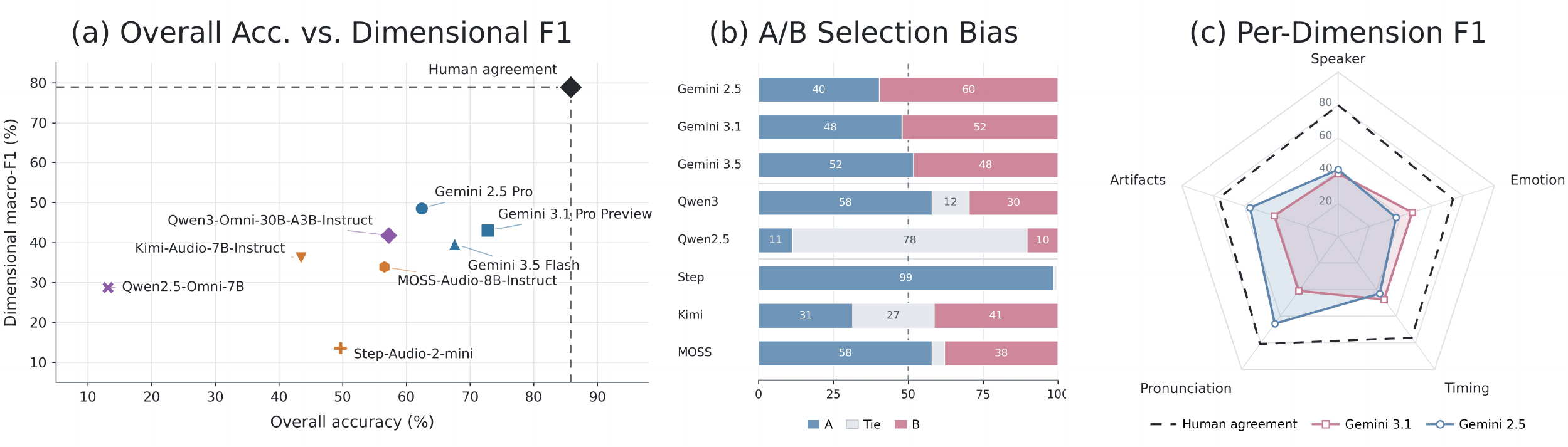}
    \caption{\textbf{Zero-shot probing} of three proprietary and five open-source audio-language models on our constructed English--Japanese test set (conducted only after all model choices were fixed.) The results reveal a clear gap to reliable human-aligned diagnostic judging.}
    \label{fig:zero_shot_diagnostic_probing}
\end{figure}

We evaluated proprietary and open-source audio-language models zero-shot on the task. All models received the same instruction. As shown in Figure~\ref{fig:zero_shot_diagnostic_probing}, three patterns emerge.
First, both Overall accuracy and dimensional macro-F1 remain below human
agreement. Second, several models exhibit strongly skewed A/B output
distributions, while Qwen2.5-Omni-7B~\citep{qwen25omni} predicts the invalid Overall
\textsc{Tie} on 78.3\% of comparisons, violating our task formulation. Third, no model performs consistently well
across all five perceptual dimensions. Off-the-shelf models are therefore not
reliable zero-shot diagnostic judges, motivating our human-calibrated
supervision and specialized judge training.

\begin{figure*}[t!]
    \centering
    \includegraphics[
        width=\textwidth,
        trim={110bp 167bp 142bp 152bp},
        clip
    ]{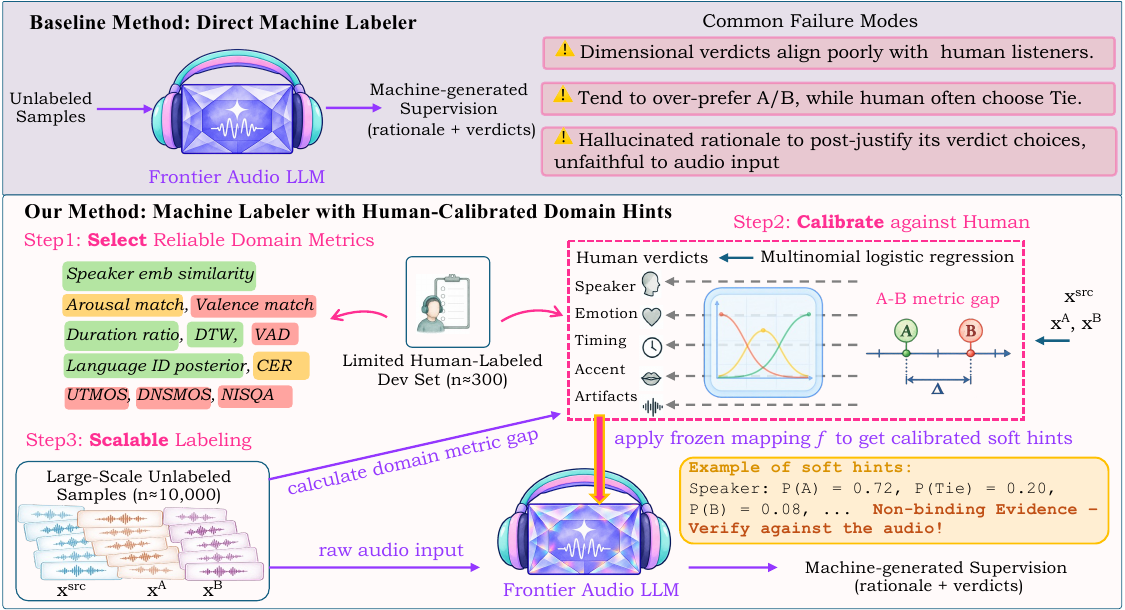}
    \caption{\textbf{Human-calibrated domain hints.} Direct machine labeling exhibits systematic disagreement with human. Our select--calibrate--scale pipeline uses limited human judgments to convert reliable domain metrics into probabilistic hints for scalable and acoustically grounded supervision.}
    \vspace{-10pt}
    \label{fig:human_calibrated_hints}
\end{figure*}

\section{Method: Scaling Supervision from Limited Human Judgments}
\label{sec:method}

Gemini-3.1-Pro~\citep{gemini31pro} achieves the highest zero-shot Overall accuracy in
Figure~\ref{fig:zero_shot_diagnostic_probing}, but its raw supervision remains
systematically misaligned with human judgments. Humans choose \textsc{Tie} for
55.1\% of dimension-level judgments, compared with only 19.8\% for Gemini, whose
rationales can also contain unsupported or post-hoc acoustic claims. We
therefore use limited human judgments to calibrate the evidence supplied to the
machine labeler through three steps: select, calibrate, and scale
(Figure~\ref{fig:human_calibrated_hints}).

\subsection{Selecting Domain Metrics}
\label{sec:select_metrics}

We first use the human-labeled development set to determine which domain
metrics provide reliable evidence for each diagnostic dimension instead of assuming they are all reliable. Let
$m_{kj}\in\mathcal{M}_k$ denote candidate measurement $j$ for dimension $k$.
For development comparison $i$, we orient each measurement so that a larger value
favors that candidate and compute its signed metric gap:
\[
    \Delta_{ik}^{(j)}
    =m_{kj}(A_i)-m_{kj}(B_i),
    \qquad
    z_{ik}\in\{A,\mathrm{Tie},B\},
\]
where $z_{ik}$ is the human-majority dimension-level verdict. We evaluate each candidate through five-fold grouped out-of-fold prediction,
keeping comparisons from the same source group in one fold. Within each fold,
standardization and a temporary probability mapping are fitted on the other
four folds, and their held-out predictions are compared with $z_{ik}$ using
macro-F1. The class-frequency baseline uses the $A$/\textsc{Tie}/$B$
frequencies estimated from the same four training folds. These fold-specific
mappings are used only to select $m_k$ and are then discarded.
Table~\ref{tab:selected_domain_metrics} summarizes the retained measurements and their out-of-fold performance. 
No candidate outperforms the baseline for Audio Artifacts, so it receives no domain hint.
Appendix~\ref{app:domain_metric_validation} reports the complete screening
results.

\begin{table}[h]
    \centering
    \caption{Retained domain metrics and grouped OOF macro-F1 on the
    human-labeled dev set.}
    \label{tab:selected_domain_metrics}
    \small
    \setlength{\tabcolsep}{3pt}
    \renewcommand{\arraystretch}{1.15}
    \begin{tabularx}{\textwidth}{
        @{}l
        >{\hsize=.85\hsize\linewidth=\hsize\centering\arraybackslash}X
        >{\hsize=.85\hsize\linewidth=\hsize\centering\arraybackslash}X
        >{\hsize=1.20\hsize\linewidth=\hsize\centering\arraybackslash}X
        >{\hsize=1.35\hsize\linewidth=\hsize\centering\arraybackslash}X
        >{\hsize=.75\hsize\linewidth=\hsize\centering\arraybackslash}X@{}
    }
        \toprule
        \textbf{Dimension}
        & \textbf{Speaker}
        & \textbf{Emotion}
        & \textbf{Timing}
        & \shortstack{\textbf{Pronunciation}}
        & \shortstack{\textbf{Artifacts}} \\
        \midrule
        \shortstack[l]{Retained\\metric(s)}
        & \shortstack{Speaker\\similarity}
        & \shortstack{Arousal\\mismatch}
        & \shortstack{Duration deviation\\+ envelope DTW}
        & \shortstack{Character error rate \\ or Language ID$^{*}$}
        & \shortstack{None} \\
        \addlinespace[2pt]
        OOF macro-F1 (\%)
        & $20.9\rightarrow29.1$
        & $19.2\rightarrow36.0$
        & $18.6\rightarrow54.7$
        & \shortstack{$26.7\rightarrow50.6$}
        & No improv. \\
        \bottomrule
    \end{tabularx}
    \parbox{0.98\textwidth}{\scriptsize
    $^{*}$OOF validation found the VoxLingua target-language posterior
    predictive only for English--Spanish, so it is used only for that
    language.}
\end{table}

\subsection{Calibrating the Selected Metrics}
\label{sec:domain_hints}

After selecting $m_k$, we refit one final mapping $f_k$ on all
development comparisons. Let $j_k^\star$ denote the retained candidate, and write
$m_k=m_{k j_k^\star}$ and
$\Delta_{ik}=\Delta_{ik}^{(j_k^\star)}$. We fit an $\ell_2$-regularized
multinomial logistic mapping
\[
    \mathbf{p}_{ik}
    =f_k(\Delta_{ik})
    =\operatorname{softmax}\!\left(
        \boldsymbol{\alpha}_k+\boldsymbol{\beta}_k
        \frac{\Delta_{ik}-\mu_k}{\sigma_k}
      \right)
    =\bigl(\Pr(A),\Pr(\mathrm{Tie}),\Pr(B)\bigr).
\]

The mapping is learned from human dimension-level verdicts using cross-entropy.
We then freeze its standardization parameters $(\mu_k,\sigma_k)$ and regression
parameters $(\boldsymbol{\alpha}_k,\boldsymbol{\beta}_k)$. The resulting
distribution represents uncertainty in the calibrated human-choice mapping
rather than forcing a hard label.

\subsection{Generating Scalable Diagnostic Supervision}
\label{sec:scalable_supervision}
\begin{table}[t]
    \centering
    \caption{
    Human-calibrated domain hints improve the dimensional alignment and \textsc{Tie}-rate calibration of Gemini~3.1~Pro as a machine labeler, with no detectable change in Overall accuracy.
    }
    \label{tab:labeling_quality}
    \fontsize{7.5pt}{9pt}\selectfont
    \setlength{\tabcolsep}{4pt}
    \renewcommand{\arraystretch}{1.05}
    \begin{tabular}{@{}lccc@{\hspace{8pt}}ccc@{}}
        \toprule
        & \multicolumn{3}{c}{Dev} & \multicolumn{3}{c}{Test} \\
        \cmidrule(lr){2-4}\cmidrule(l){5-7}
        Strategy
            & Overall Acc. \% $\uparrow$ & Dim. Macro-F1 \% $\uparrow$ & \textsc{Tie} MAE$^{*}$ $\downarrow$
            & Overall Acc. \% $\uparrow$ & Dim. Macro-F1 \% $\uparrow$ & \textsc{Tie} MAE $\downarrow$ \\
        \midrule
        Direct labeling
            & 72.6 & 44.1 & 29.0
            & \textbf{71.3} & 42.3 & 35.2 \\
        $+$ Calibrated hints
            & \textbf{75.8} & \textbf{50.9} & \textbf{22.1}
            & 69.9 & \textbf{48.5} & \textbf{24.8} \\
        \bottomrule
    \end{tabular}
    \parbox{0.99\textwidth}{\tiny
    $^{*}$The mean absolute difference between model and human \textsc{Tie} rates across dimensions; machine labelers tend to force an $A$/$B$ choice more often than humans.}
    \vspace{-10pt}
\end{table}

For each unlabeled comparison $x$, we compute the selected signed metric gap
and apply the frozen calibration mapping $f_k$:
\[
    \widehat{\mathbf{p}}_k(x)
    =f_k\!\left(m_k(A_x)-m_k(B_x)\right)
    =\bigl(\widehat{\Pr}(A),\widehat{\Pr}(\mathrm{Tie}),
    \widehat{\Pr}(B)\bigr).
\]

The machine labeler receives calibrated distributions for the
four supported dimensions alongside the raw speech, transcripts, and rubric;
Audio Artifacts is marked unknown, and no Overall hint is provided.
As illustrated in Figure~\ref{fig:human_calibrated_hints}, these hints are
non-binding: the model verifies them against the audio and may override them
before generating five dimension-level verdicts with rationales and an Overall
verdict.

Using roughly 300 human-labeled development comparisons, this pipeline guides
machine-generated verdict-and-rationale supervision for more than 10,000
unlabeled comparisons. Table~\ref{tab:labeling_quality} evaluates this
supervision before any student model is trained. On the held-out test set,
paired reference-cluster bootstrap intervals (20,000 resamples) show that
calibrated hints improve dimensional macro-F1 by 6.3 points
(95\% CI: $[3.0,9.4]$) and reduce \textsc{Tie}-rate MAE by 10.4 points
(95\% CI: $[7.4,13.4]$ reduction); both intervals exclude zero. Overall
accuracy changes by $-1.5$ points, with an interval crossing zero
(95\% CI: $[-7.0,4.0]$), indicating no detectable change.
Although \textsc{Tie} still remains under-predicted compared to human preference 
(Appendix~\ref{sec:sec:agreement}, Table~\ref{tab:tie_behavior}), the hints
make the same Gemini model a better source of machine-generated diagnostic
supervision.

\section{Training Signals for Diagnostic Speech Judges}
\label{sec:train}
Our scalable supervision specifies both verdicts and rationales, but remains imperfect because it inherits errors from the machine labeler. We therefore study not only how to generate supervision, but how a student should learn from it.
All student judges are initialized from Qwen2.5-Omni-7B; implementation details are provided in Appendix~\ref{app:training_details}.
Unless otherwise stated, results are reported as the mean
and standard deviation over three independent training runs.

\begin{table*}[t]
    \centering
    \caption{
    \textbf{Main results:}
    SFT, OPD, and RL produce distinct training signals to shape a diagnostic speech judge, rather than a
    uniform accuracy ladder.
    }
    \label{tab:main_training_comparisons}
    \fontsize{7.5pt}{9pt}\selectfont
    \definecolor{baselinepanel}{HTML}{F7E6EC}
    \definecolor{sftpanel}{HTML}{E5EFF9}
    \definecolor{opdpanel}{HTML}{F9EAD8}
    \definecolor{rlpanel}{HTML}{E5F2E7}
    \definecolor{composedpanel}{HTML}{EEE9F7}
    \newcommand{\tabresult}[2]{\ensuremath{#1\,{\color{black!65}\pm #2}}}
    \setlength{\tabcolsep}{5pt}
    \renewcommand{\arraystretch}{1.05}
    \begin{tabular*}{\textwidth}{@{\extracolsep{\fill}}p{0.38\textwidth}cccc@{}}
        \toprule
        Configuration
        & Overall Acc. (\%)
        & \shortstack{High-Consensus\\Acc. (\%)}
        & \shortstack{Low-Consensus\\Acc. (\%)}
        & \shortstack{Dim.\\Macro-F1 (\%)} \\
        \midrule
        \rowcolor{baselinepanel}\multicolumn{5}{@{}l}{\textit{Baseline}} \\
        Qwen2.5-Omni-7B (zero-shot)
            & 13.10 & 13.53 & 12.50 & 28.76 \\
        \midrule
        \rowcolor{sftpanel}\multicolumn{5}{@{}l}{\textit{A. SFT}} \\
        SFT
            & \tabresult{69.54}{0.80} & \tabresult{81.76}{1.02} & \tabresult{52.22}{3.37} & \tabresult{52.37}{0.10} \\
        \midrule
        \rowcolor{opdpanel}\multicolumn{5}{@{}l}{\textit{B. OPD: teacher choice}} \\
        Privileged: SFT teacher
            & \tabresult{71.72}{1.58} & \tabresult{85.10}{1.89} & \tabresult{52.78}{3.85} & \tabresult{52.15}{1.26} \\
        Privileged: Qwen3-30B teacher
            & \tabresult{73.79}{0.91} & \tabresult{\textbf{87.25}}{0.68} & \tabresult{54.72}{2.93} & \tabresult{44.61}{2.79} \\
        Vanilla: Qwen3-30B teacher
            & \tabresult{56.78}{1.44} & \tabresult{58.04}{3.02} & \tabresult{55.00}{0.83} & \tabresult{37.51}{0.93} \\
        \midrule
        \rowcolor{rlpanel}\multicolumn{5}{@{}l}{\textit{C. RL: reward and algorithm (initialized from SFT)}} \\
        DAPO --- Overall only
            & \tabresult{69.43}{0.87} & \tabresult{83.14}{1.89} & \tabresult{50.00}{0.83} & \tabresult{52.79}{0.83} \\
        DAPO --- Dimensional only
            & \tabresult{71.15}{1.05} & \tabresult{84.71}{1.56} & \tabresult{51.94}{0.48} & \tabresult{45.72}{1.39} \\
        DAPO --- Overall $+$ dimensional
            & \tabresult{70.57}{0.20} & \tabresult{84.12}{1.56} & \tabresult{51.39}{1.92} & \tabresult{50.19}{2.28} \\
        DAPO --- Gated dimensional reward
            & \tabresult{71.84}{1.05} & \tabresult{85.69}{1.48} & \tabresult{52.22}{0.48} & \tabresult{47.64}{1.01} \\
        GRPO --- Same gated reward
            & \tabresult{70.34}{0.91} & \tabresult{83.73}{0.34} & \tabresult{51.39}{1.73} & \tabresult{49.36}{1.64} \\
        \midrule
        \rowcolor{composedpanel}\multicolumn{5}{@{}l}{\textit{D. Composed training}} \\
        OPD initialization $+$ gated RL
            & \tabresult{\textbf{73.91}}{0.40} & \tabresult{85.88}{0.59} & \tabresult{\textbf{56.94}}{1.27} & \tabresult{\textbf{52.87}}{1.74} \\
        \bottomrule
    \end{tabular*}
\end{table*}

\paragraph{SFT establishes the diagnostic task.}

Zero-shot Qwen2.5-Omni-7B does not reliably follow the diagnostic rubric (Figure~\ref{fig:zero_shot_diagnostic_probing}): its responses often omit required verdicts or violate the binary Overall decision, yielding only 13.10\% Overall accuracy. We therefore begin with SFT on fixed verdict-and-rationale targets, which provides token-level supervision over the complete structured response. As shown in Panel A of Table~\ref{tab:main_training_comparisons}, with LoRA applied only to the language model, SFT raises Overall accuracy to 69.54\% and establishes the basic rubric following, dimensional distinctions, and output structure required by subsequent training. SFT therefore serves as the task-learning stage, although it necessarily remains exposed to errors in the fixed machine-generated targets.

\paragraph{OPD transfers the teacher's conditional judgment profile.}

Fixed-target SFT never evaluates the student on prefixes generated by its own policy. OPD instead
samples a student trajectory and minimizes token-level Jensen--Shannon
divergence from a frozen teacher along it~\cite{gkd}. Let \(p_\theta\) and \(q_\phi\)
denote the student and teacher next-token distributions. Vanilla OPD gives both
models \(x\); privileged OPD additionally gives the teacher an evidence brief
\(c\) derived from scalable supervision, while the student receives only
\(x\)~\citep{opsd}:
\[
\mathcal{L}_{\mathrm{OPD}}
=
\frac{1}{T}\sum_{t=1}^{T}
\operatorname{JSD}\!\left(
p_\theta(\cdot\mid x,y_{<t})
\,\middle\|\,
q_\phi(\cdot\mid x,c,y_{<t})
\right).
\]
Panel B of Table~\ref{tab:main_training_comparisons} shows that the teacher's input and behavior, rather than model capacity alone, determine what OPD transfers. Vanilla Qwen3~\citep{qwen3omni} supervision produces only 56.78\% Overall accuracy, whereas privileged conditioning raises it to 73.79\%. Yet the resulting dimensional macro-F1 is 44.61\%, well below the 52.15\% obtained from the SFT teacher, showing that the student inherits the teacher's judgment profile rather than a universally better judge. Holding the privileged Qwen3 teacher fixed, changing the student initialization among base, SFT, and RL checkpoints shifts Overall accuracy only within 72.07--73.79\% and yields no consistent improvement. Teacher choice and conditioning therefore dominate initialization.

\paragraph{RL improves the verdict behavior encoded by its reward.}

SFT and OPD both supervise the full token sequence, potentially transferring errors in the target or teacher rationale. Instead, we employ RL to evaluate only the parsed verdicts of each generated response, allowing the policy to improve decision behavior without treating any target rationale as ground truth. Our primary reward grants dimensional credit only after the Overall verdict is correct:
\[
R=r_{\mathrm{overall}}
+\lambda\mathbf{1}\{r_{\mathrm{overall}}=+1\}r_{\mathrm{dim}}.
\]
Here \(r_{\mathrm{overall}}\in\{-1,+1\}\), while \(r_{\mathrm{dim}}\) averages correctness over dimensions with decisive \(A/B\) targets. \textsc{Tie}-labeled dimensions and rationale text receive no direct reward.

Panel C of Table~\ref{tab:main_training_comparisons} compares Overall-only, dimension-only, additive, and gated rewards under DAPO, together with GRPO using the same gated reward. Overall-only training provides no gain, while the gated reward performs best, raising Overall accuracy from 69.54\% to 71.84\%. Nearly all of this improvement occurs on high-consensus cases (\(+3.9\) points), with no gain on low-consensus cases. At the same time, dimensional macro-F1 falls from 52.37\% to 47.64\%, and decreases further under the dimension-only reward. Because \textsc{Tie} targets receive no dimensional reward, these predictions can drift. RL therefore sharpens the decision behavior explicitly encoded by the reward rather than uniformly improving diagnostic quality.

These results show that the three signals do not form a uniform accuracy ladder: SFT establishes the task, OPD transfers the teacher's judgment profile, and RL sharpens rewarded verdict behavior. 
Additional prompt and supervision-format ablations, student-initialization controls, dual-teacher distillation, and OPD--RL interactions are consolidated in Table~\ref{tab:additional_experiments} of Appendix~\ref{app:additional_training_results}.

\section{Understanding Diagnostic Judge Behavior}
\label{sec:results}

\subsection{RQ1: How Should Scalable Supervision Be Used?}
\label{sec:rq1_specialization}

\begin{figure}[t]
\centering
\begin{minipage}[t]{0.3\linewidth}
    \centering
    \vspace{0pt}
    \includegraphics[width=\linewidth]{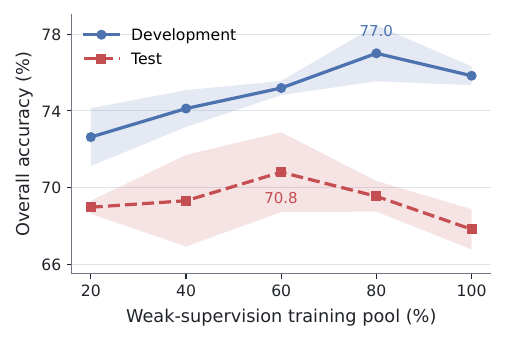}
    \caption{
    More supervision is not always helpful for SFT.
    }
    \label{fig:sft_scaling}
\end{minipage}
\hfill
\begin{minipage}[t]{0.65\linewidth}
    \centering
    \vspace{0pt}
    \captionof{table}{
    \textbf{Trainable modules for SFT.} \#M denotes trainable parameters in millions.
    }
    \label{tab:sft_trainable_modules}
    \fontsize{7.5pt}{9pt}\selectfont
    \newcommand{\sftresult}[2]{\ensuremath{#1\,{\color{black!65}\pm #2}}}
    \setlength{\tabcolsep}{3.5pt}
    \renewcommand{\arraystretch}{1.20}
    \begin{tabularx}{\linewidth}{@{}>{\raggedright\arraybackslash}Xccc@{}}
        \toprule
        Trainable modules
        & \#M
        & \shortstack{Overall\\Acc. (\%)}
        & \shortstack{Dim.\\Macro-F1 (\%)} \\
        \midrule
        Audio projector only
            & 1 & \sftresult{14.48}{1.38} & \sftresult{29.92}{0.40} \\
        LLM LoRA only
            & 323 & \sftresult{69.54}{0.80} & \sftresult{\textbf{52.37}}{0.10} \\
        LLM LoRA $+$ audio projector
            & 324 & \sftresult{70.69}{2.15} & \sftresult{52.03}{2.06} \\
        LLM LoRA $+$ audio encoder
            & 417 & \sftresult{71.95}{1.90} & \sftresult{50.79}{1.68} \\
        LLM LoRA $+$ audio encoder $+$ projector
            & 418 & \sftresult{\textbf{72.41}}{1.50} & \sftresult{52.07}{0.84} \\
        \bottomrule
    \end{tabularx}
\end{minipage}

\end{figure}

\paragraph{How much? More supervision is not always better.}
Human-calibrated domain hints improve the machine labeler, but its
development-set labels still differ from human judgments and remain imperfect
training targets (Table~\ref{tab:labeling_quality}). This makes the optimal
amount of supervision non-obvious: more comparisons increase coverage but may
also introduce less reliable targets. Figure~\ref{fig:sft_scaling} varies the
amount of machine-generated supervision used for SFT while holding the training
recipe and epochs fixed, so larger fractions receive proportionally
more updates. Performance initially improves, then declines when the full pool
is used, consistent with fixed-target SFT imitating noise or inconsistencies in
the additional comparisons. We therefore use the 80\% regime (about 8,000 comparisons)
for subsequent English--Japanese SFT comparisons and downstream experiments
initialized from SFT.

\paragraph{Where? Audio-path adaptation improves overall decisions, not dimensional diagnosis.}
Scalable supervision can be applied only to the language model, where it
primarily teaches the rubric and response structure, or also to the audio
projector and encoder, where it can adapt the speech representations used for
judging. Holding the SFT data and training recipe fixed,
Table~\ref{tab:sft_trainable_modules} shows that the audio projector alone is
insufficient, whereas extending LLM adaptation to the audio path raises Overall
accuracy from 69.54\% to 72.41\%, with most of the gain coming from the encoder.
Dimensional macro-F1 nevertheless remains unchanged, motivating audio-path
objectives designed specifically for fine-grained diagnostic perception.

\subsection{RQ2: Does Better Judgment Imply Better Diagnostic Reasoning?}
\begin{figure*}[b]
    \centering
    \includegraphics[
        width=\textwidth,
        trim=35 340 35 140,
        clip
    ]{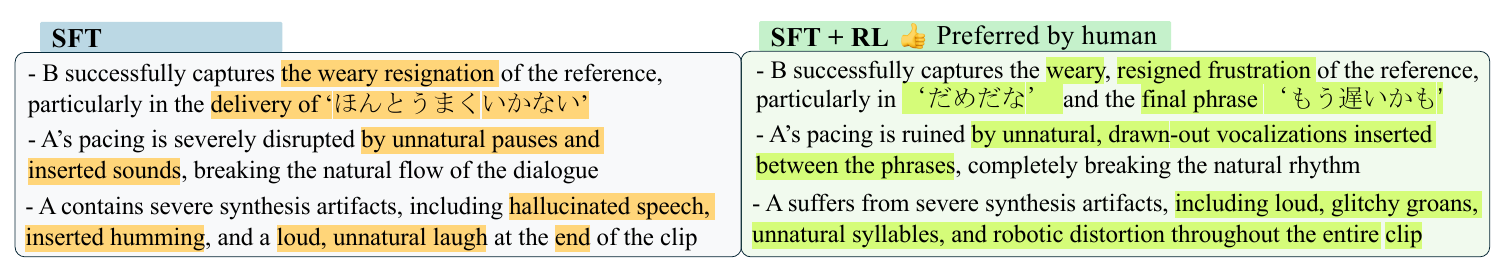}
    \caption{\textbf{RL can improve rationale grounding without direct rationale rewards.} Given identical verdicts, SFT+RL cites more specific and better-localized acoustic evidence than SFT.}
    \label{fig:rationale_example}
\end{figure*}

\paragraph{Verdict rewards can improve evidence-grounded rationales.}
A diagnostic judge must do more than predict verdicts: it should explain them
with audibly verifiable evidence localized to specific words or phrases.
We ask human listeners to compare paired SFT and SFT+RL rationales after hearing
the audios. To isolate rationale quality, we include only
cases in which both models produce the same Overall verdict. On audible-evidence grounding, listeners prefer
SFT+RL in 11 of 28 judgments, prefer SFT in one, and report no meaningful
difference in 16 (two-sided exact sign test, $p=0.006$). This suggests that
verdict rewards can improve the specificity and accuracy of cited acoustic
evidence without directly scoring rationale text. Figure~\ref{fig:rationale_example}
shows one example; Appendix~\ref{app:rationale_evaluation} provides the full
protocol and criteria. This emergent effect may arise because sequence-level
updates alter the rationale and verdict jointly, as also explored in text-only
LLMs~\citep{deepseek_r1}.

\paragraph{Persuasive Rationales Can Still Support Wrong Verdicts.}
In the main OPD comparison (Panel B of Table~\ref{tab:main_training_comparisons}), the Qwen3-teacher model achieves the highest Overall
accuracy, but its teacher is used zero-shot and has never learned our diagnostic
rubric through task-specific training. Its rationale style is also quite
different: its responses are substantially longer and more detailed than those
of the other trained judges (about 488 tokens on average, compared with 322 for SFT-teacher OPD). 
Human listeners also prefer these rationales more often than those from SFT-teacher OPD, often citing their apparent evidence grounding. Yet this advantage appears primarily when at least one dimension-level verdict is wrong or both models choose the wrong overall winner, a pattern consistent with a possible verbosity or persuasiveness bias rather than greater diagnostic reliability.
Detailed diagnostic language can therefore make an erroneous judgment appear credible: rationale fluency or persuasiveness is not diagnostic reliability.
Consistent with controlled audits showing that speech-judge rationales rarely identify the acoustic manipulation that changed a judgment~\citep{acoustic_shortcuts}, rationale quality must be evaluated against the audio and for consistency with the verdict, rather than inferred from textual plausibility alone.

\subsection{RQ3: Does \speechcritic{} Scale Across Languages?}
We evaluate cross-lingual scalability at two levels: whether the full
\speechcritic{} supervision pipeline can be instantiated for a new target
language, and whether judges trained on one or both languages transfer across
them. Starting from our primary English--Japanese setting, we repeat the supervision pipeline in Section~\ref{sec:method} for English--Spanish comparisons and train
monolingual and joint judges. On the
Spanish test set, zero-shot Qwen2.5-Omni-7B and
Gemini-3.1-Pro
reach 8.14\% and 62.87\% Overall accuracy, respectively, providing reference
points for task-specific specialization.

\begin{table*}[t]
    \centering
    \caption{
    \textbf{Cross-lingual transfer of diagnostic speech judges.}
    All models adapt
    only the LLM with LoRA. Monolingual judges transfer strongly across target
    languages, as highlighted in red.
    }
    \label{tab:multilingual_judges}
    \footnotesize
    \definecolor{jacolumn}{HTML}{EAF2FB}
    \definecolor{escolumn}{HTML}{FBEFE6}
    \definecolor{transfercell}{HTML}{F7D6DC}
    \newcommand{\multires}[2]{\ensuremath{#1\,{\color{black!65}\pm #2}}}
    \setlength{\tabcolsep}{3.5pt}
    \renewcommand{\arraystretch}{0.98}
    \begin{tabular*}{\textwidth}{@{\extracolsep{\fill}}lc
        >{\columncolor{jacolumn}}c
        >{\columncolor{jacolumn}}c
        >{\columncolor{escolumn}}c
        >{\columncolor{escolumn}}c@{}}
        \toprule
        Training
        & \shortstack{SFT data\\(comparisons)}
        & \shortstack{JA Overall\\Acc. (\%)}
        & \shortstack{JA Dim.\\Macro-F1 (\%)}
        & \shortstack{ES Overall\\Acc. (\%)}
        & \shortstack{ES Dim.\\Macro-F1 (\%)} \\
        \midrule
        JA only                & 7,840
            & \multires{69.54}{0.80} & \multires{52.37}{0.10}
            & \cellcolor{transfercell}\multires{65.91}{2.17} & \cellcolor{transfercell}\multires{48.23}{2.03} \\
        ES only                & 8,000
            & \cellcolor{transfercell}\multires{70.92}{0.80} & \cellcolor{transfercell}\multires{50.52}{1.17}
            & \multires{66.02}{1.32} & \multires{50.00}{0.89} \\
        JA+ES combined         & 15,840
            & \multires{71.38}{1.72} & \multires{50.10}{1.30}
            & \multires{66.67}{1.35} & \multires{49.75}{0.56} \\
        \bottomrule
    \end{tabular*}
\end{table*}

Table~\ref{tab:multilingual_judges} shows that each monolingual judge transfers
strongly to the other language, while joint training incurs no clear
loss relative to either monolingual model. Thus, both the supervision framework
and the resulting diagnostic behavior scale across the two target languages
studied, and the latter can be consolidated into a single judge.
Appendix~\ref{app:crosslingual_evaluation} further reports results on the public VOX-DUB benchmark~\citep{voxdub}. Because its dimensions and protocol differ from ours, we treat it as an external reference rather than a directly comparable benchmark.

\section{Conclusion}
We introduced \speechcritic{}, a framework for learning diagnostic, acoustically grounded speech judges from limited human preferences. Its \textit{select--calibrate--scale} pipeline identifies reliable acoustic evidence, maps it to uncertainty-aware human-choice probabilities, and uses the frozen mappings to guide a machine labeler that also listens to the original audio. This approach scales roughly 300 human-labeled comparisons into more than 10,000 verdict-and-rationale training comparisons, improving agreement with human dimension-level judgments and better reflecting when listeners consider two candidates tied.

Our results show that constructing scalable supervision and learning from it jointly shape a judge's decisions and explanations. SFT establishes task competence, although more supervision is not always better and audio-path adaptation does not improve every diagnostic criterion. OPD transfers the teacher's judgment profile, while RL sharpens the verdict behavior encoded by its reward. Human evaluation shows that RL can also improve rationales' grounding in specific, localized acoustic cues, even without directly rewarding rationale text. Yet persuasive rationales can still accompany incorrect verdicts, underscoring the need to evaluate verdict correctness and acoustic grounding separately, rather than treating fluent explanations as evidence of reliable judgment. Both the supervision pipeline and the resulting judges also transfer across Japanese and Spanish. Together, these findings demonstrate a path from limited human preferences to diagnostic speech judges.

\bibliography{iclr2027_conference}

@inproceedings{audiojudge,
    title = "{A}udio{J}udge: Understanding What Works in Large Audio Model Based Speech Evaluation",
    author = "Manakul, Potsawee  and
      Gan, Woody Haosheng  and
      Ryan, Michael J  and
      Khan, Ali Sartaz  and
      Sirichotedumrong, Warit  and
      Pipatanakul, Kunat  and
      Held, William Barr  and
      Yang, Diyi",
    editor = "Demberg, Vera  and
      Inui, Kentaro  and
      Marquez, Llu{\'i}s",
    booktitle = "Proceedings of the 19th Conference of the {E}uropean Chapter of the {A}ssociation for {C}omputational {L}inguistics (Volume 1: Long Papers)",
    month = mar,
    year = "2026",
    address = "Rabat, Morocco",
    publisher = "Association for Computational Linguistics",
    url = "https://aclanthology.org/2026.eacl-long.168/",
    doi = "10.18653/v1/2026.eacl-long.168",
    pages = "3644--3663",
    ISBN = "979-8-89176-380-7"
}

@inproceedings{trace,
  title={Hearing between the lines: Unlocking the reasoning power of LLMs for speech evaluation},
  author={Chandra, Arjun and Miller, Kevin and Ravichandran, Venkatesh and Papayiannis, Constantinos and Saligrama, Venkatesh},
  booktitle={Findings of the Association for Computational Linguistics: EACL 2026},
  pages={2895--2916},
  year={2026}
}

@misc{jastin,
      title={{JASTIN}: Aligning LLMs for Zero-Shot Audio and Speech Evaluation via Natural Language Instructions}, 
      author={Leying Zhang and Bowen Shi and Haibin Wu and Bach Viet Do and Yanmin Qian},
      year={2026},
      eprint={2605.04505},
      archivePrefix={arXiv},
      primaryClass={eess.AS},
      url={https://arxiv.org/abs/2605.04505}, 
}

@inproceedings{qualispeech,
  title     = {{QualiSpeech}: A Speech Quality Assessment Dataset with Natural Language Reasoning and Descriptions},
  author    = {Wang, Siyin and Yu, Wenyi and Chen, Xianzhao and Tian, Xiaohai and Zhang, Jun and Lu, Lu and Tsao, Yu and Yamagishi, Junichi and Wang, Yuxuan and Zhang, Chao},
  booktitle = {Proceedings of the 63rd Annual Meeting of the Association for Computational Linguistics (Volume 1: Long Papers)},
  month     = jul,
  year      = {2025},
  address   = {Vienna, Austria},
  publisher = {Association for Computational Linguistics},
  pages     = {23588--23609},
  doi       = {10.18653/v1/2025.acl-long.1150},
  url       = {https://aclanthology.org/2025.acl-long.1150/}
}

@inproceedings{sqllm,
    title = "{S}peech{LLM}-as-Judges: Towards General and Interpretable Speech Quality Evaluation",
    author = "Wang, Hui  and
      Zhao, Jinghua  and
      Yang, Yifan  and
      Liu, Shujie  and
      Chen, Junyang  and
      Zhang, Yanzhe  and
      Zhao, Shiwan  and
      Li, Jinyu  and
      Zhou, Jiaming  and
      Sun, Haoqin  and
      Lu, Yan  and
      Qin, Yong",
    editor = "Liakata, Maria  and
      Moreira, Viviane P.  and
      Zhang, Jiajun  and
      Jurgens, David",
    booktitle = "Proceedings of the 64th Annual Meeting of the {A}ssociation for {C}omputational {L}inguistics (Volume 1: Long Papers)",
    month = jul,
    year = "2026",
    address = "San Diego, California, United States",
    publisher = "Association for Computational Linguistics",
    url = "https://aclanthology.org/2026.acl-long.349/",
    doi = "10.18653/v1/2026.acl-long.349",
    pages = "7675--7700",
    ISBN = "979-8-89176-390-6"}

@article{wavreward,
  title={Wavreward: Spoken dialogue models with generalist reward evaluators},
  author={Ji, Shengpeng and Liang, Tianle and Li, Yangzhuo and Zuo, Jialong and Fang, Minghui and He, Jinzheng and Chen, Yifu and Liu, Zhengqing and Jiang, Ziyue and Cheng, Xize and others},
  journal={arXiv preprint arXiv:2505.09558},
  year={2025}
}

@article{speechjudge,
  title={{SpeechJudge}: Towards Human-Level Judgment for Speech Naturalness},
  author={Zhang, Xueyao and Wang, Chaoren and Liao, Huan and Li, Ziniu and Wang, Yuancheng and Wang, Li and Jia, Dongya and Chen, Yuanzhe and Li, Xiulin and Chen, Zhuo and Wu, Zhizheng},
  journal={arXiv preprint arXiv:2511.07931},
  year={2025}
}

@inproceedings{unisrm,
  title     = {{UniSRM}: A Unified Speech Reward Model for Reasoning-Based Fine-grained Assessment},
  author    = {Wang, Yuanyuan and Yang, Dongchao and Deng, Yayue and Wu, Zhiyong and Guo, Steven Y. and Meng, Helen M. and Wu, Xixin},
  booktitle = {Proceedings of the 64th Annual Meeting of the Association for Computational Linguistics (Volume 1: Long Papers)},
  month     = jul,
  year      = {2026},
  address   = {San Diego, California, United States},
  publisher = {Association for Computational Linguistics},
  pages     = {46346--46366},
  doi       = {10.18653/v1/2026.acl-long.2150},
  url       = {https://aclanthology.org/2026.acl-long.2150/}
}

@misc{gsrm,
      title={{GSRM}: Generative Speech Reward Model for Speech RLHF}, 
      author={Maohao Shen and Tejas Jayashankar and Osama Hanna and Naoyuki Kanda and Yancheng Wang and Kateřina Žmolíková and Ruiming Xie and Niko Moritz and Anfeng Xu and Yashesh Gaur and Gregory Wornell and Qing He and Jilong Wu},
      year={2026},
      eprint={2602.13891},
      archivePrefix={arXiv},
      primaryClass={cs.SD},
      url={https://arxiv.org/abs/2602.13891}, 
}

@inproceedings{prometheus,
  title={Prometheus: Inducing fine-grained evaluation capability in language models},
  author={Kim, Seungone and Shin, Jay and Jang, Joel and Longpre, Shayne and Lee, Hwaran and Yun, Sangdoo and Shin, Ryan and Kim, Sungdong and Thorne, James and Seo, Minjoon and others},
  booktitle={International Conference on Learning Representations},
  volume={2024},
  pages={29927--29962},
  year={2024}
}

@article{judgelm,
      title={JudgeLM: Fine-tuned Large Language Models are Scalable Judges}, 
      author={Lianghui Zhu and Xinggang Wang and Xinlong Wang},
      year={2023},
      eprint={2310.17631},
      archivePrefix={arXiv},
      primaryClass={cs.CL}
}

@inproceedings{autoj,
  title={Generative judge for evaluating alignment},
  author={Li, Junlong and Sun, Shichao and Yuan, Weizhe and Fan, Run-Ze and Liu, Pengfei and others},
  booktitle={International Conference on Learning Representations},
  volume={2024},
  pages={27547--27574},
  year={2024}
}

@article{cloud,
  title={Critique-out-loud reward models},
  author={Ankner, Zachary and Paul, Mansheej and Cui, Brandon and Chang, Jonathan D and Ammanabrolu, Prithviraj},
  journal={arXiv preprint arXiv:2408.11791},
  year={2024}
}

@article{rlaif,
  title={Rlaif vs. rlhf: Scaling reinforcement learning from human feedback with ai feedback},
  author={Lee, Harrison and Phatale, Samrat and Mansoor, Hassan and Mesnard, Thomas and Ferret, Johan and Lu, Kellie and Bishop, Colton and Hall, Ethan and Carbune, Victor and Rastogi, Abhinav and others},
  journal={arXiv preprint arXiv:2309.00267},
  year={2023}
}

@article{constitutional_ai,
  title={Constitutional ai: Harmlessness from ai feedback},
  author={Bai, Yuntao and Kadavath, Saurav and Kundu, Sandipan and Askell, Amanda and Kernion, Jackson and Jones, Andy and Chen, Anna and Goldie, Anna and Mirhoseini, Azalia and McKinnon, Cameron and others},
  journal={arXiv preprint arXiv:2212.08073},
  year={2022}
}

@misc{self_taught_evaluator,
      title={Self-Taught Evaluators}, 
      author={Tianlu Wang and Ilia Kulikov and Olga Golovneva and Ping Yu and Weizhe Yuan and Jane Dwivedi-Yu and Richard Yuanzhe Pang and Maryam Fazel-Zarandi and Jason Weston and Xian Li},
      year={2024},
      eprint={2408.02666},
      archivePrefix={arXiv},
      primaryClass={cs.CL},
      url={https://arxiv.org/abs/2408.02666}, 
}

@article{snorkel,
   title={Snorkel: rapid training data creation with weak supervision},
   volume={11},
   ISSN={2150-8097},
   url={http://dx.doi.org/10.14778/3157794.3157797},
   DOI={10.14778/3157794.3157797},
   number={3},
   journal={Proceedings of the VLDB Endowment},
   publisher={Association for Computing Machinery (ACM)},
   author={Ratner, Alexander and Bach, Stephen H. and Ehrenberg, Henry and Fries, Jason and Wu, Sen and Ré, Christopher},
   year={2017},
   month=Nov, pages={269–282} }

@article{deepseek_r1,
  title   = {{DeepSeek-R1}: Incentivizing Reasoning Capability in {LLM}s via Reinforcement Learning},
  author  = {{DeepSeek-AI}},
  journal = {Nature},
  volume  = {645},
  pages   = {633--638},
  year    = {2025},
  doi     = {10.1038/s41586-025-09422-z}
}

@misc{qwen25omni,
      title={Qwen2.5-Omni Technical Report}, 
      author={Jin Xu and Zhifang Guo and Jinzheng He and Hangrui Hu and Ting He and Shuai Bai and Keqin Chen and Jialin Wang and Yang Fan and Kai Dang and Bin Zhang and Xiong Wang and Yunfei Chu and Junyang Lin},
      year={2025},
      eprint={2503.20215},
      archivePrefix={arXiv},
      primaryClass={cs.CL},
      url={https://arxiv.org/abs/2503.20215}, 
}

@misc{qwen3omni,
      title={Qwen3-Omni Technical Report}, 
      author={Jin Xu and Zhifang Guo and Hangrui Hu and Yunfei Chu and Xiong Wang and Jinzheng He and Yuxuan Wang and Xian Shi and Ting He and Xinfa Zhu and Yuanjun Lv and Yongqi Wang and Dake Guo and He Wang and Linhan Ma and Pei Zhang and Xinyu Zhang and Hongkun Hao and Zishan Guo and Baosong Yang and Bin Zhang and Ziyang Ma and Xipin Wei and Shuai Bai and Keqin Chen and Xuejing Liu and Peng Wang and Mingkun Yang and Dayiheng Liu and Xingzhang Ren and Bo Zheng and Rui Men and Fan Zhou and Bowen Yu and Jianxin Yang and Le Yu and Jingren Zhou and Junyang Lin},
      year={2025},
      eprint={2509.17765},
      archivePrefix={arXiv},
      primaryClass={cs.CL},
      url={https://arxiv.org/abs/2509.17765}, 
}

@misc{gemini31pro,
  author       = {{Google}},
  title        = {{Gemini 3.1 Pro Preview}},
  year         = {2026},
  howpublished = {Gemini API model documentation},
  url          = {https://ai.google.dev/gemini-api/docs/models/gemini-3.1-pro-preview}
}

@misc{gemini25,
  title         = {{Gemini 2.5}: Pushing the Frontier with Advanced Reasoning,
                   Multimodality, Long Context, and Next Generation Agentic Capabilities},
  author        = {{Gemini Team}},
  year          = {2025},
  eprint        = {2507.06261},
  archiveprefix = {arXiv},
  primaryclass  = {cs.AI},
  doi           = {10.48550/arXiv.2507.06261},
}

@misc{gemini35flash,
  author       = {{Google DeepMind}},
  title        = {{Gemini 3.5 Flash} Model Card},
  year         = {2026},
  howpublished = {Model card},
  url          = {https://deepmind.google/models/model-cards/gemini-3-5-flash/}
}

@misc{stepaudio2,
      title={Step-Audio 2 Technical Report}, 
      author={Boyong Wu and Chao Yan and Chen Hu and Cheng Yi and Chengli Feng and Fei Tian and Feiyu Shen and Gang Yu and Haoyang Zhang and Jingbei Li and Mingrui Chen and Peng Liu and Wang You and Xiangyu Tony Zhang and Xingyuan Li and Xuerui Yang and Yayue Deng and Yechang Huang and Yuxin Li and Yuxin Zhang and Zhao You and Brian Li and Changyi Wan and Hanpeng Hu and Jiangjie Zhen and Siyu Chen and Song Yuan and Xuelin Zhang and Yimin Jiang and Yu Zhou and Yuxiang Yang and Bingxin Li and Buyun Ma and Changhe Song and Dongqing Pang and Guoqiang Hu and Haiyang Sun and Kang An and Na Wang and Shuli Gao and Wei Ji and Wen Li and Wen Sun and Xuan Wen and Yong Ren and Yuankai Ma and Yufan Lu and Bin Wang and Bo Li and Changxin Miao and Che Liu and Chen Xu and Dapeng Shi and Dingyuan Hu and Donghang Wu and Enle Liu and Guanzhe Huang and Gulin Yan and Han Zhang and Hao Nie and Haonan Jia and Hongyu Zhou and Jianjian Sun and Jiaoren Wu and Jie Wu and Jie Yang and Jin Yang and Junzhe Lin and Kaixiang Li and Lei Yang and Liying Shi and Li Zhou and Longlong Gu and Ming Li and Mingliang Li and Mingxiao Li and Nan Wu and Qi Han and Qinyuan Tan and Shaoliang Pang and Shengjie Fan and Siqi Liu and Tiancheng Cao and Wanying Lu and Wenqing He and Wuxun Xie and Xu Zhao and Xueqi Li and Yanbo Yu and Yang Yang and Yi Liu and Yifan Lu and Yilei Wang and Yuanhao Ding and Yuanwei Liang and Yuanwei Lu and Yuchu Luo and Yuhe Yin and Yumeng Zhan and Yuxiang Zhang and Zidong Yang and Zixin Zhang and Binxing Jiao and Daxin Jiang and Heung-Yeung Shum and Jiansheng Chen and Jing Li and Xiangyu Zhang and Yibo Zhu},
      year={2025},
      eprint={2507.16632},
      archivePrefix={arXiv},
      primaryClass={cs.CL},
      url={https://arxiv.org/abs/2507.16632}, 
}

@article{kimiaudio,
  title={Kimi-audio technical report},
  author={Ding, Ding and Ju, Zeqian and Leng, Yichong and Liu, Songxiang and Liu, Tong and Shang, Zeyu and Shen, Kai and Song, Wei and Tan, Xu and Tang, Heyi and others},
  journal={arXiv preprint arXiv:2504.18425},
  year={2025}
}

@misc{mossaudio,
      title={MOSS-Audio Technical Report}, 
      author={Chen Yang and Chufan Yu and Hanfu Chen and Jie Zhu and Jingqi Chen and Ke Chen and Wenxuan Wang and Yang Wang and Yaozhou Jiang and Yi Jiang and Zhengyuan Lin and Ziqi Chen and Zhaoye Fei and Chenghao Liu and Donghua Yu and Jun Zhan and Kang Yu and Kexin Huang and Liwei Fan and Mingshu Chen and Qinyuan Cheng and Ruixiao Li and Shimin Li and Songlin Wang and Xingjian Zhao and Yang Gao and Yitian Gong and Yiyang Zhang and Zhe Xu and Xipeng Qiu},
      year={2026},
      eprint={2606.01802},
      archivePrefix={arXiv},
      primaryClass={cs.SD},
      url={https://arxiv.org/abs/2606.01802}, 
}

@misc{ms_swift,
      title={{SWIFT}:A Scalable lightWeight Infrastructure for Fine-Tuning}, 
      author={Yuze Zhao and Jintao Huang and Jinghan Hu and Xingjun Wang and Yunlin Mao and Daoze Zhang and Hong Zhang and Zeyinzi Jiang and Zhikai Wu and Baole Ai and Ang Wang and Wenmeng Zhou and Yingda Chen},
      year={2025},
      eprint={2408.05517},
      archivePrefix={arXiv},
      primaryClass={cs.CL},
      url={https://arxiv.org/abs/2408.05517}, 
}

@misc{dapo,
      title={{DAPO}: An Open-Source LLM Reinforcement Learning System at Scale}, 
      author={Qiying Yu and Zheng Zhang and Ruofei Zhu and Yufeng Yuan and Xiaochen Zuo and Yu Yue and Weinan Dai and Tiantian Fan and Gaohong Liu and Lingjun Liu and Xin Liu and Haibin Lin and Zhiqi Lin and Bole Ma and Guangming Sheng and Yuxuan Tong and Chi Zhang and Mofan Zhang and Wang Zhang and Hang Zhu and Jinhua Zhu and Jiaze Chen and Jiangjie Chen and Chengyi Wang and Hongli Yu and Yuxuan Song and Xiangpeng Wei and Hao Zhou and Jingjing Liu and Wei-Ying Ma and Ya-Qin Zhang and Lin Yan and Mu Qiao and Yonghui Wu and Mingxuan Wang},
      year={2025},
      eprint={2503.14476},
      archivePrefix={arXiv},
      primaryClass={cs.LG},
      url={https://arxiv.org/abs/2503.14476}, 
}

@misc{deepseekmath,
      title={DeepSeekMath: Pushing the Limits of Mathematical Reasoning in Open Language Models}, 
      author={Zhihong Shao and Peiyi Wang and Qihao Zhu and Runxin Xu and Junxiao Song and Xiao Bi and Haowei Zhang and Mingchuan Zhang and Y. K. Li and Y. Wu and Daya Guo},
      year={2024},
      eprint={2402.03300},
      archivePrefix={arXiv},
      primaryClass={cs.CL},
      url={https://arxiv.org/abs/2402.03300}, 
}

@inproceedings{gkd,
  title={On-policy distillation of language models: Learning from self-generated mistakes},
  author={Agarwal, Rishabh and Vieillard, Nino and Zhou, Yongchao and Stanczyk, Piotr and Ramos Garea, Sabela and Geist, Matthieu and Bachem, Olivier},
  booktitle={International Conference on Learning Representations},
  volume={2024},
  pages={21246--21263},
  year={2024}
}

@article{opsd,
  title={Self-Distilled Reasoner: On-Policy Self-Distillation for Large Language Models},
  author={Zhao, Siyan and Xie, Zhihui and Liu, Mengchen and Huang, Jing and Pang, Guan and Chen, Feiyu and Grover, Aditya},
  journal={arXiv preprint arXiv:2601.18734},
  year={2026}
}

@article{lora,
  title={Lora: Low-rank adaptation of large language models},
  author={Hu, Edward J and Shen, Yelong and Wallis, Phillip and Allen-Zhu, Zeyuan and Li, Yuanzhi and Wang, Shean and Wang, Lu and Chen, Weizhu},
  journal={arXiv preprint arXiv:2106.09685},
  year={2021}
}

@misc{voxdub,
  title        = {{VOX-DUB}: a new benchmark that puts {AI} dubbing to the test},
  author       = {{Toloka team}},
  howpublished = {\url{https://toloka.ai/blog/ai-dubbing-benchmark/}},
  year         = {2025},
  month        = sep # "~9"
}

@INPROCEEDINGS{wespeaker,
  author={Wang, Hongji and Liang, Chengdong and Wang, Shuai and Chen, Zhengyang and Zhang, Binbin and Xiang, Xu and Deng, Yanlei and Qian, Yanmin},
  booktitle={ICASSP 2023 - 2023 IEEE International Conference on Acoustics, Speech and Signal Processing (ICASSP)}, 
  title={Wespeaker: A Research and Production Oriented Speaker Embedding Learning Toolkit}, 
  year={2023},
  volume={},
  number={},
  pages={1-5},
  doi={10.1109/ICASSP49357.2023.10096626}}

@ARTICLE{wagner2023emotion,
  author={Wagner, Johannes and Triantafyllopoulos, Andreas and Wierstorf, Hagen and Schmitt, Maximilian and Burkhardt, Felix and Eyben, Florian and Schuller, Björn W.},
  journal={IEEE Transactions on Pattern Analysis and Machine Intelligence}, 
  title={Dawn of the Transformer Era in Speech Emotion Recognition: Closing the Valence Gap}, 
  year={2023},
  volume={45},
  number={9},
  pages={10745-10759},
  doi={10.1109/TPAMI.2023.3263585}}

@inproceedings{emotion2vec,
  title     = {{emotion2vec}: Self-Supervised Pre-Training for Speech Emotion Representation},
  author    = {Ma, Ziyang and Zheng, Zhisheng and Ye, Jiaxin and Li, Jinchao and Gao, Zhifu and Zhang, Shiliang and Chen, Xie},
  booktitle = {Findings of the Association for Computational Linguistics: ACL 2024},
  year      = {2024},
  pages     = {15747--15760},
  doi       = {10.18653/v1/2024.findings-acl.931}
}

@article{sakoe1978dtw,
  title   = {Dynamic Programming Algorithm Optimization for Spoken Word Recognition},
  author  = {Sakoe, Hiroaki and Chiba, Seibi},
  journal = {IEEE Transactions on Acoustics, Speech, and Signal Processing},
  volume  = {26},
  number  = {1},
  pages   = {43--49},
  year    = {1978},
  doi     = {10.1109/TASSP.1978.1163055}
}

@misc{silero_vad,
  author = {Silero Team},
  title = {Silero VAD: pre-trained enterprise-grade Voice Activity Detector (VAD), Number Detector and Language Classifier},
  year = {2024},
  publisher = {GitHub},
  journal = {GitHub repository},
  howpublished = {\url{https://github.com/snakers4/silero-vad}},
  commit = {insert_some_commit_here},
  email = {hello@silero.ai}
}

@misc{qwen3_asr,
      title={Qwen3-ASR Technical Report}, 
      author={Xian Shi and Xiong Wang and Zhifang Guo and Yongqi Wang and Pei Zhang and Xinyu Zhang and Zishan Guo and Hongkun Hao and Yu Xi and Baosong Yang and Jin Xu and Jingren Zhou and Junyang Lin},
      year={2026},
      eprint={2601.21337},
      archivePrefix={arXiv},
      primaryClass={cs.CL},
      url={https://arxiv.org/abs/2601.21337}, 
}

@INPROCEEDINGS{voxlingua107,
  author={Valk, Jörgen and Alumäe, Tanel},
  booktitle={2021 IEEE Spoken Language Technology Workshop (SLT)}, 
  title={{VOXLINGUA107}: A Dataset for Spoken Language Recognition}, 
  year={2021},
  volume={},
  number={},
  pages={652-658},
  doi={10.1109/SLT48900.2021.9383459}}

@misc{utmos,
      title={{UTMOS}: UTokyo-SaruLab System for VoiceMOS Challenge 2022}, 
      author={Takaaki Saeki and Detai Xin and Wataru Nakata and Tomoki Koriyama and Shinnosuke Takamichi and Hiroshi Saruwatari},
      year={2022},
      eprint={2204.02152},
      archivePrefix={arXiv},
      primaryClass={cs.SD},
      url={https://arxiv.org/abs/2204.02152}, 
}

@INPROCEEDINGS{dnsmos,
  author={Reddy, Chandan K A and Gopal, Vishak and Cutler, Ross},
  booktitle={ICASSP 2021 - 2021 IEEE International Conference on Acoustics, Speech and Signal Processing (ICASSP)}, 
  title={{DNSMOS}: A Non-Intrusive Perceptual Objective Speech Quality Metric to Evaluate Noise Suppressors}, 
  year={2021},
  volume={},
  number={},
  pages={6493-6497},
  doi={10.1109/ICASSP39728.2021.9414878}}

@inproceedings{nisqa,
   title={{NISQA}: A Deep CNN-Self-Attention Model for Multidimensional Speech Quality Prediction with Crowdsourced Datasets},
   url={http://dx.doi.org/10.21437/Interspeech.2021-299},
   DOI={10.21437/interspeech.2021-299},
   booktitle={Interspeech 2021},
   publisher={ISCA},
   author={Mittag, Gabriel and Naderi, Babak and Chehadi, Assmaa and Möller, Sebastian},
   year={2021},
   month=Aug, pages={2127–2131},
   collection={interspeech_2021} }

@article{qwen3tts,
  title={Qwen3-tts technical report},
  author={Hu, Hangrui and Zhu, Xinfa and He, Ting and Guo, Dake and Zhang, Bin and Wang, Xiong and Guo, Zhifang and Jiang, Ziyue and Hao, Hongkun and Guo, Zishan and others},
  journal={arXiv preprint arXiv:2601.15621},
  year={2026}
}

@article{cosyvoice2,
  title={Cosyvoice 2: Scalable streaming speech synthesis with large language models},
  author={Du, Zhihao and Wang, Yuxuan and Chen, Qian and Shi, Xian and Lv, Xiang and Zhao, Tianyu and Gao, Zhifu and Yang, Yexin and Gao, Changfeng and Wang, Hui and others},
  journal={arXiv preprint arXiv:2412.10117},
  year={2024}
}

@misc{acoustic_shortcuts,
  title={{Louder, Longer, Livelier}: Acoustic Shortcuts and Underspecified Rationales in Speech {LLM} Judges},
  author={Huo, Mingyue and Mehta, Shivam and Jawade, Bhavin and Lan, Yinghong and Li, Haoqi},
  year={2026},
  note={Preprint}
}
\bibliographystyle{iclr2027_conference}

\appendix
\section{Human Annotation and Benchmark Analysis}
\label{app:human_agreement}

\subsection{Annotation Protocol and Prompt}

Each annotation page presented an English reference utterance and two
target-language candidate utterances, denoted A and B. Raters first selected the
better candidate
overall, with no overall \textsc{Tie} option. They then compared the candidates along the
five dimensions defined in Section~\ref{sec:sec:task_formulation}: Speaker,
Emotion, Timing, Pronunciation/Accent, and Audio Artifacts. Each dimension-level
question allowed $A$, \textsc{Tie}, or $B$, and the page additionally
provided an optional comment field. The deployed interfaces were bilingual. We
reproduce the English instructions below, replacing the specific target-language
name with [\emph{target language}] so that the protocol applies to both of
our language settings.

\begin{promptbox}{Instructions for Human Preference Annotation}
\small
\begin{itemize}
    \setlength{\itemsep}{2pt}
    \setlength{\topsep}{4pt}
    \item First listen to the English reference. Then compare the two
    [\emph{target language}] speech candidates and choose which is better
    overall.
    \item After choosing the Overall winner, judge the five dimensions below independently. A
    candidate may win Overall while losing or tying on an individual dimension.
    \item Use \textsc{Tie} only for the five dimension-level judgments when there is no clear difference.
    Overall requires $A$ or $B$.
    \item Do not evaluate translation accuracy or whether the
    [\emph{target language}] text is a literal translation of the English.
    Both candidates use the same target text. Focus only on audible speech
    characteristics and technical audio quality.
    \item Comments are optional.
\end{itemize}

\textbf{Overall verdict.}
Using the English utterance as the reference, which [\emph{target language}]
speech candidate is better overall? Choose one winner.

\textbf{Dimension definitions.}
\begin{description}
    \item[D1 Speaker.] Match the reference speaker's vocal
    timbre, pitch range, vocal weight, and speaking style.
    \item[D2 Emotion.] Match the type and intensity of emotion in the English
    reference; do not simply reward more emotion.
    \item[D3 Timing.] Pauses, duration, speaking rate, rhythm, and
    natural [\emph{target language}] delivery all match the English reference.
    \item[D4 Pronunciation/Accent.] Clear, correct, authentic
    [\emph{target language}] articulation and pronunciation, without
    unnatural foreign-accent leakage.
    \item[D5 Audio Artifacts.] Only noise, clipping, cutoffs, glitches,
    missing/extra sounds, or audio artifacts. Do not penalize emotion,
    timing, or pronunciation again here.
\end{description}
\end{promptbox}

\subsection{Annotation Results and Benchmark Composition}

Table~\ref{tab:human_benchmark_statistics} reports only the development and test
comparisons used in our experiments. Most received three independent annotations.
To assess whether three-person majorities remain stable with a larger panel, we
also collected 20 annotations for 30 shared anchor comparisons. These anchors are
included in the benchmark totals but not in the high- and low-consensus
counts, which refer strictly to $3$--$0$ and $2$--$1$ outcomes among the
standard three-rater comparisons. Comparisons without a strict majority for Overall are
excluded from the benchmark.

\begin{table}[t]
    \centering
    \small
    \setlength{\tabcolsep}{6pt}
    \caption{Human benchmark used in our experiments. Counts include only
    development and test comparisons with a strict majority for Overall. Consensus
    counts are computed among standard three-rater comparisons: high-consensus is
    $3$--$0$ and low-consensus is $2$--$1$.}
    \label{tab:human_benchmark_statistics}
    \begin{tabular}{lrrrr}
        \toprule
        & \multicolumn{2}{c}{\textbf{English--Japanese}}
        & \multicolumn{2}{c}{\textbf{English--Spanish}} \\
        \cmidrule(lr){2-3}\cmidrule(lr){4-5}
        & \textbf{Dev} & \textbf{Test} & \textbf{Dev} & \textbf{Test} \\
        \midrule
        Comparisons & 315 & 290 & 307 & 307 \\
        High-consensus & 180 & 160 & 139 & 136 \\
        Low-consensus & 122 & 116 & 153 & 156 \\
        \bottomrule
    \end{tabular}
\end{table}

\subsection{Do the Five Dimensions Explain Overall Preference?}
\label{app:rubric_composition}

We test whether the five dimensions form a coherent account of holistic human
preference rather than an arbitrary diagnostic checklist.  For comparison $i$ and
dimension $d$, let $c^{A}_{id}$ and $c^{B}_{id}$ be the numbers of votes for A
and B among $R_i$ raters, and define the signed preference score
\begin{equation}
    z_{id}=\frac{c^{A}_{id}-c^{B}_{id}}{R_i},
    \qquad s_i=\sum_{d=1}^{5} z_{id}.
\end{equation}
A dimension-level \textsc{Tie} contributes zero to $z_{id}$.  For the primary
check, we fit a one-feature logistic calibration,
$P(y_i=A\mid s_i)=\sigma(\alpha+\beta s_i)$, where $y_i$ is the human Overall
majority label.  On the 302 standard three-rater English--Japanese development
comparisons, five-fold out-of-fold evaluation keeps comparisons derived from the same
source group in the same fold and gives 93.0\%
accuracy, 0.209 log loss, and a 0.059 Brier score.  A more flexible logistic
model with dimension-specific preference and \textsc{Tie}-rate features performs no
better (92.1\% accuracy, 0.228 log loss, and 0.065 Brier score).  The same
equal-weight analysis reaches 82.6\% accuracy on 293 English--Spanish
development comparisons, whose annotations exhibit lower agreement.  These results
show that the five perceptual judgments form a coherent rubric: together, they
capture most holistic choices without requiring a learned dimension hierarchy.
They do not establish that the dimensions are exhaustive or that
Overall preference is causally determined by their equal-weight sum.

\subsection{Agreement Analysis}
\label{sec:sec:agreement}
\paragraph{Why report human agreement?}
Multi-dimensional speech-quality judgments are intrinsically subjective: two attentive listeners
can hear the same candidates yet disagree about which difference should
determine the Overall verdict or whether a perceptual difference is large
enough to avoid a \textsc{Tie}. We therefore report a human agreement reference
to contextualize model accuracy, not as a strict upper bound on achievable
performance. A model near 70\% accuracy should be interpreted differently when
human labels themselves exhibit substantial disagreement than when the task has
nearly deterministic labels.

\paragraph{Primary estimator: single rater versus panel gold.}
For comparison $i$, let $c_i(y)$ denote the number of panel votes for label $y$, let
$R_i$ be the number of raters, and let $g_i$ be the panel-majority gold label.
Our reference accuracy is
\begin{equation}
    \frac{1}{N}\sum_{i=1}^{N}\frac{c_i(g_i)}{R_i},
\end{equation}
which is the expected accuracy of drawing one panel rater uniformly at random
and scoring that judgment against the panel gold. Every comparison receives equal
weight. For dimension-level judgments, we form the corresponding expected confusion
matrix over comparisons with a unique panel-majority label, matching the model
evaluation, compute three-class macro-F1 for each dimension, and average across
the five dimensions.
\textsc{Tie} recall analogously averages the fraction of raters choosing \textsc{Tie} on
dimensions whose panel gold is \textsc{Tie}. For consensus-stratified human
agreement, we use only standard three-rater comparisons: agreement is 100\% on
high-consensus comparisons and $66.67\%$ on low-consensus comparisons; the
separate 20-rater anchors are excluded from these two estimates.

This estimator is useful because it has the same interpretation and metric
scale as the reported model scores. It is mildly optimistic---the sampled rater
also contributes to the majority label---so we call it a human agreement
\emph{reference}, rather than an independent-rater ceiling. On the
English--Japanese test set it yields 85.82\% Overall accuracy, 78.84\%
dimensional macro-F1, and 86.07\% dimensional \textsc{Tie} recall. These values quantify
the subjectivity of the benchmark; they do not excuse model errors or imply
that human disagreement is irreducible.

\paragraph{\textsc{Tie} behavior on the constructed test set.}
Table~\ref{tab:tie_behavior} characterizes how often each system uses
\textsc{Tie} on all 290 comparisons in the English--Japanese test set. The relatively
high human-majority \textsc{Tie} prevalence describes the composition of this benchmark,
not a universal property of speech evaluation. Direct Gemini is substantially
more decisive than the human panel; calibrated Gemini moves its \textsc{Tie} profile
toward the human distribution, and the OPD judge moves closer still overall,
although the effect is strongly dimension-dependent. In particular, the OPD
judge closely matches Pronunciation/Accent (D4) and Audio Artifacts (D5) but
remains highly decisive on Emotion (D2) and Timing (D3).
Matching marginal \textsc{Tie} prevalence also does not imply comparison-level correctness,
which is why dimensional macro-F1 remains the primary diagnostic metric.
Here, calibrated Gemini uses human-calibrated domain hints on the 272
synthesized-candidate comparisons and the audio-only policy on the 18
natural-versus-synthesized controls.
Accordingly, Table~\ref{tab:labeling_quality} evaluates the labeling-policy
comparison on the locked 272-comparison primary subset, whereas the
dimension-wise rates below describe all 290 comparisons; its aggregate
\textsc{Tie} MAE therefore cannot be reconstructed by averaging the full-set
rates below.

\begin{table*}[t]
    \centering
    \caption{
    \textbf{Dimensional \textsc{Tie} prevalence on our constructed English--Japanese test set.}
    Human values are the percentages of comparisons whose panel-majority label is
    \textsc{Tie}; model values are the percentages predicted as \textsc{Tie}.
    The OPD judge uses an SFT-trained teacher; its results are averaged over
    three seeds.
    }
    \label{tab:tie_behavior}
    \small
    \setlength{\tabcolsep}{6pt}
    \renewcommand{\arraystretch}{1.08}
    \begin{tabularx}{\textwidth}{@{}Xcccc@{}}
        \toprule
        Dimension
            & \shortstack{Human-majority\\\textsc{Tie} (\%)}
            & \shortstack{Direct Gemini\\\textsc{Tie} (\%)}
            & \shortstack{Calibrated Gemini\\\textsc{Tie} (\%)}
            & \shortstack{OPD judge\\\textsc{Tie} (\%)} \\
        \midrule
        D1: Speaker       & 52.03 &  9.59 & 18.08 & 34.32 \\
        D2: Emotion  & 25.38 &  3.08 &  4.62 &  4.36 \\
        D3: Timing   & 39.63 &  7.41 & 12.59 &  4.57 \\
        D4: Pronunciation and accent        & 70.42 & 21.48 & 46.83 & 72.89 \\
        D5: Audio artifacts     & 87.89 & 57.44 & 64.71 & 87.77 \\
        \bottomrule
    \end{tabularx}
\end{table*}

\paragraph{Alternative agreement summaries.}
Raw pairwise agreement measures how often two raters select the same label, and
Krippendorff's $\alpha$ additionally corrects for agreement expected from the
label marginals. Both are valuable descriptions of annotation reliability, but
neither is directly comparable to model accuracy or macro-F1. Leave-one-rater-out
agreement is more independent in principle, but is severely downward biased
for three-rater panels: removing one majority voter from a low-consensus comparison
leaves a $1$--$1$ tie among the remaining raters. Our 20-rater anchors permit a less
biased leave-one-out diagnostic, but they are too few to represent the full test
distribution. We therefore use single-rater-versus-gold agreement in the main
table and treat pairwise, chance-corrected, and large-panel estimates as
complementary reliability analyses.

\section{Data Construction and Human-Calibrated Domain Hints}
\label{app:data_construction}

\subsection{Constructing Diverse Speech Comparisons}

\paragraph{Candidate construction.}
For data construction, we partition the in-house reference utterances
into training, development, and test sets using an 8:1:1 split. All
utterances from the same speaker are assigned to the same split, preventing
closely related speech from crossing evaluation boundaries.
We use publicly available synthesis systems, including
Qwen3-TTS-12Hz-1.7B-Base and CosyVoice2-0.5B
\citep{qwen3tts,cosyvoice2}, to construct candidate pools with diverse
perceptual characteristics and failure modes. For Speaker, we perform voice
conversion using same-speaker and different-speaker conditioning references,
creating variation in the preservation of speaker characteristics. For
Pronunciation/Accent, we contrast target-language voice cloning with
cross-lingual conditioning, which can introduce foreign-accent leakage.
For Emotion, Timing, and Audio Artifacts, we sample 12 outputs from each of
two multilingual TTS architectures. This multi-output pool captures both
within-model stochastic variation and systematic differences between
synthesis families.
Within each target dimension, candidates are paired randomly to form the
speech comparisons. Domain measurements are used to characterize and audit the resulting pool, not to select the winner within a comparison.
We then deterministically randomize their assignment to
positions A and B and balance the retained pool across target dimensions and
candidate positions. This procedure produces comparisons ranging from clear
to subtle while reducing shortcuts based on candidate position or synthesis
system.

\paragraph{Human benchmark and scalable pool.}
From the development and test splits, we sample dimension-balanced subsets for
human annotation. 
We also add a small natural--synthesized control set to test
whether the judge can compare candidates with different provenance rather than
only candidates produced by synthesis systems. These controls are a secondary
test rather than the primary benchmark construction. 
The resulting benchmark composition is reported in
Table~\ref{tab:human_benchmark_statistics}.  Human development judgments are
used to select and calibrate candidate metrics; development-set Overall
accuracy separately selects trained checkpoints as described in
Appendix~\ref{app:training_details}. Test judgments remain untouched until
final evaluation. The much larger training split receives the
resulting soft hints and machine-generated verdict-and-rationale supervision.
Thus, the construction procedure creates diverse comparisons, human labels
determine which measurements deserve trust, and the held-out test set plays no
role in either decision.

\subsection{Selecting Domain Metrics for Human-Calibrated Hints}
\label{app:domain_metric_validation}

This section provides the metric-selection evidence behind the calibration step
in Section~\ref{sec:domain_hints}; the retained mappings are subsequently used
for scalable supervision as described in Section~\ref{sec:scalable_supervision}.
We screened 14 prespecified candidate metrics using grouped out-of-fold macro-F1 against a class-frequency baseline, supplemented by small expert listening pilots when automatic validation was inconclusive; the strongest supported metric for each dimension was retained. Table~\ref{tab:domain_metric_validation} summarizes the candidate measurements we examined.

\begin{table*}[t]
    \centering
    \caption{Candidate domain metrics, human validation, and use in the final
    soft hints. OOF denotes grouped out-of-fold evaluation against human
    development-set verdicts; pilot listening studies were conducted by
    linguist experts. \textbf{Retained} metrics enter the probability mapping,
    \emph{auxiliary} metrics support construction or auditing, and rejected
    metrics are not exposed to the machine labeler.}
    \label{tab:domain_metric_validation}
    \small
    \setlength{\tabcolsep}{6pt}
    \renewcommand{\arraystretch}{1.08}
    \begin{tabularx}{\textwidth}{@{}p{2.25cm}p{5.4cm}X@{}}
        \toprule
        Dimension & Metric implementation & Human validation and use \\
        \midrule
        \multirow{2}{2.25cm}{Speaker}
        & WeSpeaker \texttt{w2vbert2\_mfa} embedding cosine to the
          target-language reference~\citep{wespeaker}
        & \textbf{Retained.}  OOF macro-F1
          improves from 20.9 to 29.1. \\
        & The same WeSpeaker embedding cosine to the reference
        & Rejected as the calibrated feature; retained only as a sanity check. \\
        \midrule
        \multirow{4}{2.25cm}{Emotion}
        & Arousal mismatch from audEERING wav2vec2 MSP-DIM~\citep{wagner2023emotion}
        & \textbf{Retained.} Strongest screened emotion signal; OOF macro-F1
          improves from 19.2 to 36.0. \\
        & Valence mismatch from the same MSP-DIM model
        & Rejected; weaker than arousal in feature screening. \\
        & \texttt{emotion2vec+ large} embedding similarity~\citep{emotion2vec}
        & Rejected; unreliable on target-language speech in expert listening. \\
        & F0/pitch-contour match
        & Rejected; no improvement over arousal in a 30-pair expert pilot. \\
        \midrule
        \multirow{2}{2.25cm}{Timing}
        & Composite duration--envelope score: absolute reference-to-candidate
          duration-ratio deviation plus scaled amplitude-envelope dynamic time
          warping (DTW)~\citep{sakoe1978dtw}
        & \textbf{Retained.} OOF macro-F1 improves from 18.6 to 54.7; adding
          envelope DTW improves an expert pilot from 66.7\% to 74.1\% on 27 pairs. \\
        & Silero-VAD speech/pause alignment~\citep{silero_vad}
        & Rejected; no improvement over duration in expert pilots. \\
        \midrule
        \multirow{2}{2.25cm}{Pronunciation/Accent}
        & Qwen3-ASR-1.7B character error rate (CER)~\citep{qwen3_asr}
        & \textbf{Retained for Japanese.} OOF macro-F1 improves from 26.7 to
          50.6; unsuitable for phonetic Spanish inputs. \\
        & VoxLingua107 ECAPA target-language posterior~\citep{voxlingua107}
        & \textbf{Retained for Spanish}; auxiliary for Japanese pool auditing. \\
        \midrule
        \multirow{3}{2.25cm}{Audio Artifacts}
        & UTMOS22-Strong~\citep{utmos}
        & Rejected; no OOF improvement over the class-frequency baseline. \\
        & DNSMOS P.835 (SIG, BAK, OVRL)~\citep{dnsmos}
        & Rejected; no consistent improvement over UTMOS in a 29-pair expert pilot. \\
        & NISQA-TTS~\citep{nisqa}
        & Rejected; below chance in the same expert pilot. \\
        \bottomrule
    \end{tabularx}
\end{table*}

\section{Training and Evaluation Setup}
\label{app:training_details}

\subsection{Training Objectives and Implementation Details}
\label{app:training_implementation}

Unless stated otherwise, the student is Qwen2.5-Omni-7B, computation uses
bfloat16, the audio encoder and projector are frozen, and LoRA~\citep{lora} is applied to all
linear layers of the language model. We implement SFT and RL with
ms-swift~\citep{ms_swift} and
OPD with a custom PyTorch/PEFT trainer. The principal SFT and RL experiments use
eight NVIDIA A100 80\,GB GPUs, while OPD uses eight NVIDIA H200 GPUs. Checkpoints
are selected by development-set Overall accuracy. All reported
three-seed comparisons use seeds 42, 123, and 456.

The English--Japanese machine-labeling stage covers 11,824 comparisons. After
format and confidence filtering, 9,797 enter the SFT scaling pool; percentage
conditions use rounded subset sizes, so the 80\% setting contains 7,840
comparisons. The English--Spanish release separately contains 8,000
comparisons.

\paragraph{Supervised fine-tuning (SFT).}
We train with the ms-swift autoregressive SFT implementation on fixed
machine-generated responses containing the dimension-level verdicts, rationales,
and Overall verdict. The principal setting uses 7,840 comparisons (80\% of the
scalable supervision pool) and trains for six epochs; every data-size condition
uses the same number of epochs. LoRA has rank 128, scaling factor 256, and
dropout 0.05. We use a per-device batch size of 2 and four gradient-accumulation
steps, giving an effective batch size of 64 across eight GPUs. The learning
rate is $5\times10^{-5}$ with cosine decay and 5\% warmup; weight decay is zero.
The maximum sequence length is 5,120 tokens, and training uses DeepSpeed
ZeRO-2. The audio-path ablations retain this recipe while extending LoRA to the
4.6-million-parameter audio-to-LLM projector, to the audio tower's attention
and feed-forward layers, or to both. The remaining audio-tower parameters stay
frozen.

\paragraph{On-policy distillation (OPD).}
For each training input, the student greedily generates a trajectory of at most
1,024 tokens. The frozen teacher is then evaluated on the same trajectory, and
we minimize the mean full-vocabulary Jensen--Shannon divergence between their
next-token distributions at every generated position.
Vanilla OPD gives both models the ordinary judge input, whereas privileged OPD
additionally gives the teacher the evidence brief associated with that comparison.
The student is initialized either from Qwen2.5-Omni-7B or from SFT-trained
weights. Teachers are a frozen base or task-adapted Qwen2.5-Omni-7B, or
Qwen3-Omni-30B-A3B-Instruct. The principal runs use 7,840 comparisons and LoRA with
rank 128, scaling factor 256, and dropout 0.05 on the language-model query, key,
value, output, gate, up, and down projections. AdamW uses learning rate
$5\times10^{-5}$ and weight decay 0.01. The per-device batch size is 1 with
eight accumulation steps, giving an effective batch size of 64 across eight
GPUs. We train for at most 750 optimizer steps.

\paragraph{Reinforcement learning (RL).}
We train with DAPO~\citep{dapo} as implemented in ms-swift and compute rewards directly from
parsed verdicts; no learned reward model or rationale reward is used. Controlled
reward comparisons contain 651 labeled prompts and
initialize from SFT- or OPD-trained weights. LoRA has rank 64, scaling factor
128, and dropout 0.05. The Overall reward is $+1$ for the correct $A$/$B$ verdict
and $-1$ for an incorrect, missing,
unparseable, or \textsc{Tie} Overall verdict. The dimensional reward is the mean $\pm1$
score over dimensions whose reference verdict is $A$ or $B$; reference \textsc{Tie} cases
are excluded. We compare Overall-only, dimension-only, additive, and
Overall-gated dimensional rewards, using dimensional weight $\lambda=0.3$ for
the primary gated configuration; we separately evaluate GRPO~\citep{deepseekmath} as an algorithmic
control. We sample eight completions per prompt at temperature 1.0, capping each
at 2,048 tokens and the full sequence at 4,096 tokens.
The learning rate is $5\times10^{-6}$, the per-device batch size is 1, gradient
accumulation is 32, and training runs for at most 300 optimizer steps.

\subsection{Code Release}
\label{app:code_release}

We release configuration-driven training code for SFT, OPD, and RL. SFT and
RL build on ms-swift~\citep{ms_swift}, while OPD uses a custom distributed
PyTorch/Transformers/PEFT implementation inspired by OPSD~\citep{opsd}. Users
can launch each stage through \texttt{train/run.sh} after specifying the data,
model or adapter, and output paths in the provided YAML configurations. The
release includes an example JSONL manifest showing the required audio paths,
transcripts, verdicts, and rationales. Trained checkpoints are not included;
users can initialize from the specified public base models or compatible
checkpoints.

\subsection{Evaluation Metrics and Statistical Protocol}
\label{app:evaluation_protocol}

An output is successfully parsed only when its Overall verdict and all five
dimension-level verdicts can be extracted; otherwise it is counted as incorrect.
Because the Overall gold label is binary, a predicted \textsc{Tie} is also incorrect.
For each dimension, comparisons without a strict human-vote majority are excluded
from that dimension's F1 calculation. Three-class macro-F1 is computed on the
remaining comparisons and then averaged
equally across the five dimensions. For consensus-stratified model results,
high consensus means a majority share of at least 0.8, and low consensus means
a strict majority below 0.8. This rater-count-agnostic rule places the 14
twenty-rater English--Japanese test anchors into 10 high- and four
low-consensus comparisons; together with the standard three-rater counts in
Table~\ref{tab:human_benchmark_statistics}, the reported model slices therefore
contain 170 and 120 comparisons, respectively. Reported seed variation is the sample
standard deviation across independently trained runs, not a confidence
interval. Where a paired-bootstrap interval is explicitly reported, we
jointly resample reference clusters for both systems over 20,000 bootstrap
replicates and report the percentile 95\% interval of their metric difference.

\clearpage
\subsection{Language-Agnostic Judge System Prompt}
\label{app:judge_prompt}

We use the following system prompt for the language-agnostic diagnostic judge.

\begin{promptbox}[left=6pt,right=6pt,top=4pt,bottom=4pt,before skip=4pt,after skip=0pt]{System Prompt for the Diagnostic Speech Judge}
\tiny
\setlength{\parskip}{0pt}
You are an expert diagnostic evaluator of reference-conditioned cross-lingual
speech comparisons. You receive three audio clips:
\begin{itemize}
    \setlength{\itemsep}{1pt}
    \setlength{\topsep}{3pt}
    \item \textbf{Reference}: speech in the source language;
    \item \textbf{Candidate A}: a speech candidate in the target language; and
    \item \textbf{Candidate B}: a speech candidate in the target language.
\end{itemize}
You may also receive the source-language reference transcript and the
target-language transcript. Compare Candidates A and B based on what you hear
and return the Overall verdict.

\medskip
\textbf{Important constraints.}
\begin{itemize}
    \setlength{\itemsep}{1pt}
    \setlength{\topsep}{3pt}
    \item Do not invent numerical measurements, timestamps, or
    signal-processing facts.
    \item Do not judge translation meaning. Assume both candidates express the
    intended content unless an audible speech error is clearly present.
    \item Use transcripts only to localize audible evidence, such as a word,
    phrase, pause, emphasis, emotional turn, breath, laugh, or the beginning or
    end of the utterance.
    \item Prefer concrete audible observations over generic claims. For
    example, describe the audible speaker characteristics rather than merely
    saying that they match.
    \item Select \texttt{[[A]]} or \texttt{[[B]]} only when there is a concrete
    audible difference. Otherwise select \texttt{[[tie]]} and briefly explain why.
    \item Judge each dimension independently before deriving the Overall verdict.
    Include a ``\textsc{Tie} note'' only when the dimension-level verdict is \texttt{[[tie]]}.
\end{itemize}

\textbf{Conflicting evidence.}
Describe trade-offs honestly: one candidate may be better on some dimensions
and worse on others. Base the Overall verdict on the most consequential audible
differences rather than counting dimension-level winners. When the trade-off is
subtle, state the uncertainty while still returning the required Overall $A$/$B$
verdict.

\medskip
\textbf{Dimension definitions.}

\textbf{D1 Speaker.}
Use the reference as the anchor. When audible, first characterize its vocal
timbre, pitch range, vocal weight, and speaking style; then identify which
candidate better preserves these attributes. Exact cross-language voice
identity is not required. Overall similarity in audible speaker characteristics
is more important than identical timbre. Localize the observation to a phrase
or change in delivery when possible.

\textbf{D2 Emotion.}
Use the reference as the anchor. Compare emotional type and intensity, pitch
movement, stress, hesitation, warmth, urgency, restraint, and emotional arc.
Matching may mean remaining calm when the reference is calm, rather than being
more expressive. Name the audible emotion or attitude when possible. Keep
emotion mismatch or limited expressiveness here; pronunciation and accent belong to D4.

\textbf{D3 Timing.}
Use the reference as the anchor while also considering absolute naturalness.
Compare rhythm, speaking rate, pause placement, duration fit, breath timing, and
the naturalness of the delivery. Do not reward rushed or stretched speech merely
because its duration matches the reference. Localize observations to words,
phrase boundaries, pauses, breaths, or utterance boundaries.

\textbf{D4 Pronunciation/Accent.}
Judge primarily from the candidates. Listen for clear target-language
articulation, natural phoneme realization, source-language accent leakage, or
pronunciation that sounds non-native or strained. Do not penalize natural
regional pronunciation or favor one regional standard. Separate pronunciation
and accent from breathy or nasal voice quality, emotional stiffness, and
audio artifacts. Localize issues to words, syllables, phones, or
phrase-level intonation when possible.

\textbf{D5 Audio Artifacts.}
Judge primarily from the candidates and consider only technical defects:
background noise or hiss; clipping, cutoff, dropped endings, or missing audio;
glitches, clicks, pops, discontinuities, or unstable or distorted audio; unexpected
extra sounds; and robotic or metallic artifacts that affect the
signal itself. Do not use D5 for speaker characteristics, emotion, timing,
pronunciation, or accent. If neither candidate has clear
audio artifacts, select \texttt{[[tie]]}. Do not treat context-appropriate room
tone or reverberation as a defect. Localize any defect to where it occurs.

\medskip
\textbf{Required output format.}

\textbf{[Reference Anchor]}
\begin{itemize}
    \setlength{\itemsep}{1pt}
    \setlength{\topsep}{3pt}
    \item Describe the speaker profile.
    \item Describe the emotion or intent.
    \item Describe salient pacing, pauses, breaths, or other audible cues.
    \item Give a transcript-localized cue useful for comparing the candidates.
\end{itemize}

For each dimension, use:

\textbf{[Dimension Name]}\\
Candidate A: \texttt{<audible observation>}\\
Candidate B: \texttt{<audible observation>}\\
\textsc{Tie} note: \texttt{<include only for [[tie]]>}\\
Dimension-level verdict: \texttt{[[A]] / [[B]] / [[tie]]}\\[3pt]
\textbf{[Overall Verdict]}\\
Summarize the main audible reasons for the Overall verdict in one to three
sentences. Explicitly mention important trade-offs.\\
Overall verdict: \texttt{[[A]] / [[B]]}
\end{promptbox}

\section{Additional Experimental Results}
\label{app:additional_training_results}
\label{app:opd_initialization}

Table~\ref{tab:additional_experiments} consolidates secondary baselines and
controls omitted from the main text.

\paragraph{A. Zero-shot baselines.}
The zero-shot comparison covers proprietary and open audio-language models of
different sizes. Their Overall and dimensional performance varies widely, and
no model provides consistently balanced diagnostic behavior without
task-specific training. We evaluate the Gemini~\citep{gemini25,gemini35flash},
Qwen Omni~\citep{qwen25omni,qwen3omni}, Step-Audio~\citep{stepaudio2},
Kimi-Audio~\citep{kimiaudio}, and MOSS-Audio~\citep{mossaudio} model families.

\paragraph{B. Amount of scalable supervision.}
Holding the SFT recipe and number of epochs fixed, performance improves over
the smaller fractions but declines when the full machine-generated supervision pool is used.
This non-monotonic trend confirms that more imperfect supervision is not
necessarily better.

\paragraph{C. Target and input controls.}
These targeted ablations vary whether SFT receives rationales, a reference
summary, transcripts, or language-parametrized instructions. The results show
that performance does not hinge on any single textual field; the additional
natural-speech control also checks that the learned comparison is not confined
to pairs of synthesized candidates. Because these are single-run controls, we
treat their differences as sensitivity evidence rather than model rankings.

\paragraph{D. OPD conditioning and initialization.}
Privileged task information changes Qwen3 from an ineffective teacher into the
strongest teacher for Overall accuracy, whereas changing the student
initialization has a smaller effect. Updating the audio encoder can shift the
Overall--dimensional balance, but does not remove the teacher-specific profile
transferred by OPD.

\paragraph{E. Dual teachers and target language.}
The task-adapted Qwen2.5 teacher better preserves the dimensional rubric,
whereas Qwen3 provides stronger Overall-verdict supervision, motivating us
to test whether their signals are complementary. \emph{MeanBlend} averages the
two next-token distributions before distillation; \emph{CorrectGated} uses only
teachers whose Overall verdict matches the training target, falling back to the
Qwen2.5 teacher if neither does; and \emph{SumJSD} computes a separate
distillation loss for each teacher and averages the two losses. MeanBlend is
evaluated over three seeds, while the other two are targeted single-run
controls. None uniformly dominates the stronger single-teacher configurations;
the Spanish rows further test the two teacher choices on a second target
language.

\paragraph{F. RL and composed training.}
Reward definition and RL algorithm both change which behavior improves:
verdict-gated RL gives the largest Overall gain from the LLM-only SFT
initialization, whereas dimension-only rewards incur the largest loss in
dimensional macro-F1. Audio-path adaptation and gated RL each improve Overall
accuracy separately, but their combination reaches 71.03\%, below the
audio-adapted SFT model (72.41\%) and the LLM-only SFT model followed by RL
(71.84\%). In the seed with complete comparison-level predictions, the two
interventions correct only five of the same errors (Jaccard overlap $=0.13$), so
the non-additivity is not explained by redundant corrections alone. Starting
RL from OPD gives the strongest composed result, while updating the encoder
during this stage offers no consistent additional gain.

\paragraph{Test-time self-consistency.}
For five representative judges, we also sample 20 responses for each test comparison at
temperature 0.7 and aggregate the dimension-level verdicts by majority vote (one
training seed per judge). Voting usually improves over an average sampled
response but does not reliably outperform greedy decoding; a paired bootstrap
finds a significant gain in Overall accuracy only for SFT, while the strongest OPD
judge slightly declines. Agreement across draws is more useful as a confidence
signal (correctness AUC 0.58--0.68), suggesting that test-time sampling exposes
uncertainty more reliably than it adds diagnostic capability.

\begin{table}[p]
\centering
\caption{\textbf{Additional experimental results and controls.}
Unless marked ES, results use the English--Japanese test set. Three-run results
are mean $\pm$ standard deviation across training seeds; single-run results are
zero-shot evaluations or targeted ablations. Trained checkpoints are selected
by development-set Overall accuracy.}
\label{tab:additional_experiments}
\begingroup
\scriptsize
\definecolor{baselinepanel}{HTML}{F7E6EC}
\definecolor{sftpanel}{HTML}{E5EFF9}
\definecolor{opdpanel}{HTML}{F9EAD8}
\definecolor{rlpanel}{HTML}{E5F2E7}
\newcommand{\extraresult}[2]{\ensuremath{#1\,{\color{black!65}\pm #2}}}
\setlength{\tabcolsep}{2.5pt}
\renewcommand{\arraystretch}{1.04}
\begin{tabular*}{\textwidth}{@{\extracolsep{\fill}}p{0.40\textwidth}ccrrrr@{}}
\toprule
Configuration & Eval. & Runs
& \shortstack{Overall\\Acc. (\%)}
& \shortstack{High-Consensus\\Acc. (\%)}
& \shortstack{Low-Consensus\\Acc. (\%)}
& \shortstack{Dim.\\Macro-F1 (\%)} \\
\midrule

\rowcolor{baselinepanel}
\multicolumn{7}{@{}l}{\textit{A. Zero-shot audio-language models}} \\
Gemini-2.5-Pro                    & JA & 1 & 62.41 & 67.65 & 55.00 & 48.58 \\
Gemini-3.5-Flash                  & JA & 1 & 67.59 & 74.12 & 58.33 & 39.67 \\
\addlinespace[1pt]
Qwen3-Omni-30B                    & JA & 1 & 57.24 & 61.18 & 51.67 & 41.75 \\
Qwen2.5-Omni-7B                   & JA & 1 & 13.10 & 13.53 & 12.50 & 28.76 \\
\addlinespace[1pt]
Step-Audio-2-mini                 & JA & 1 & 49.66 & 50.00 & 49.17 & 13.58 \\
Kimi-Audio-7B                     & JA & 1 & 43.45 & 37.65 & 51.67 & 36.11 \\
MOSS-Audio-8B                     & JA & 1 & 56.55 & 63.53 & 46.67 & 33.87 \\
\midrule

\rowcolor{sftpanel}
\multicolumn{7}{@{}l}{\textit{B. SFT: amount of scalable supervision}} \\
20\% of the supervision pool      & JA & 3 & \extraresult{68.97}{0.34} & \extraresult{79.61}{1.48} & \extraresult{53.89}{1.73} & \extraresult{49.45}{2.75} \\
40\% of the supervision pool      & JA & 3 & \extraresult{69.31}{2.39} & \extraresult{80.78}{4.34} & \extraresult{53.06}{1.27} & \extraresult{47.92}{2.52} \\
60\% of the supervision pool      & JA & 3 & \extraresult{70.80}{2.08} & \extraresult{82.55}{1.48} & \extraresult{54.17}{3.00} & \extraresult{51.95}{1.09} \\
80\% of the supervision pool      & JA & 3 & \extraresult{69.54}{0.80} & \extraresult{81.76}{1.02} & \extraresult{52.22}{3.37} & \extraresult{52.37}{0.10} \\
100\% of the supervision pool     & JA & 3 & \extraresult{67.82}{1.05} & \extraresult{81.57}{0.90} & \extraresult{48.33}{1.44} & \extraresult{51.14}{0.44} \\
\midrule

\rowcolor{sftpanel}
\multicolumn{7}{@{}l}{\textit{C. SFT: target and input ablations}} \\
Full verdict-and-rationale supervision & JA & 1 & 70.69 & 84.12 & 51.67 & 51.49 \\
Verdicts only                           & JA & 1 & 70.69 & 84.12 & 51.67 & 52.96 \\
Without the reference summary          & JA & 1 & 74.14 & 84.71 & 59.17 & 48.93 \\
Without transcripts                    & JA & 1 & 74.14 & 86.47 & 56.67 & 52.45 \\
Language-parametrized instructions     & JA & 1 & 72.41 & 85.88 & 53.33 & 53.74 \\
Additional natural-speech controls     & JA & 1 & 72.07 & 83.53 & 55.83 & 47.63 \\
\midrule

\rowcolor{opdpanel}
\multicolumn{7}{@{}l}{\textit{D. OPD: teacher conditioning and student initialization}} \\
Vanilla OPD; audio-adapted SFT teacher; LLM only
    & JA & 3 & \extraresult{70.11}{0.80} & \extraresult{81.96}{2.23} & \extraresult{53.33}{2.50} & \extraresult{48.83}{1.53} \\
Vanilla OPD; Qwen3 teacher; LLM only
    & JA & 3 & \extraresult{56.78}{1.44} & \extraresult{58.04}{3.02} & \extraresult{55.00}{0.83} & \extraresult{37.51}{0.93} \\
Vanilla OPD; Qwen3 teacher; LLM $+$ encoder
    & JA & 3 & \extraresult{54.83}{1.50} & \extraresult{57.06}{2.35} & \extraresult{51.67}{1.44} & \extraresult{37.63}{3.78} \\
\addlinespace[1pt]
Privileged OPD; base student; LLM only
    & JA & 3 & \extraresult{73.79}{0.91} & \extraresult{87.25}{0.68} & \extraresult{54.72}{2.93} & \extraresult{44.61}{2.79} \\
Privileged OPD; SFT student; LLM only
    & JA & 3 & \extraresult{72.99}{0.40} & \extraresult{85.29}{1.56} & \extraresult{55.56}{3.15} & \extraresult{45.53}{0.45} \\
Privileged OPD; RL student; LLM only
    & JA & 3 & \extraresult{72.07}{0.34} & \extraresult{85.49}{0.90} & \extraresult{53.06}{0.96} & \extraresult{43.32}{1.47} \\
Privileged OPD; base student; LLM $+$ encoder
    & JA & 3 & \extraresult{74.14}{2.60} & \extraresult{86.47}{1.02} & \extraresult{56.67}{5.00} & \extraresult{47.21}{1.89} \\
Privileged OPD; audio-adapted SFT student; LLM $+$ encoder
    & JA & 3 & \extraresult{71.38}{1.58} & \extraresult{85.29}{1.02} & \extraresult{51.67}{2.89} & \extraresult{45.98}{1.16} \\
\midrule

\rowcolor{opdpanel}
\multicolumn{7}{@{}l}{\textit{E. OPD: dual teachers and language settings}} \\
Dual teacher (MeanBlend); LLM $+$ encoder
    & JA & 3 & \extraresult{72.07}{1.50} & \extraresult{86.27}{1.22} & \extraresult{51.94}{2.55} & \extraresult{52.74}{6.40} \\
Dual teacher (CorrectGated); LLM $+$ encoder
    & JA & 1 & 71.03 & 85.29 & 50.83 & 49.53 \\
Dual teacher (SumJSD); LLM $+$ encoder
    & JA & 1 & 70.34 & 85.88 & 48.33 & 56.39 \\
\addlinespace[1pt]
English--Spanish OPD; SFT teacher
    & ES & 1 & 68.40 & 76.47 & 61.99 & 47.30 \\
English--Spanish OPD; Qwen3 teacher
    & ES & 1 & 67.10 & 80.15 & 56.73 & 45.62 \\
\midrule

\rowcolor{rlpanel}
\multicolumn{7}{@{}l}{\textit{F. RL: reward, algorithm, and composed-training controls}} \\
LLM-only SFT
    & JA & 3 & \extraresult{69.54}{0.80} & \extraresult{81.76}{1.02} & \extraresult{52.22}{3.37} & \extraresult{52.37}{0.10} \\
LLM-only SFT $+$ gated RL
    & JA & 3 & \extraresult{71.84}{1.05} & \extraresult{85.69}{1.48} & \extraresult{52.22}{0.48} & \extraresult{47.64}{1.01} \\
Audio-adapted SFT
    & JA & 3 & \extraresult{72.41}{1.50} & \extraresult{83.53}{0.59} & \extraresult{56.67}{4.33} & \extraresult{52.07}{0.84} \\
Audio-adapted SFT $+$ gated RL
    & JA & 3 & \extraresult{71.03}{1.38} & \extraresult{84.90}{1.22} & \extraresult{51.39}{1.73} & \extraresult{51.71}{0.25} \\
\addlinespace[1pt]
LLM-only SFT $+$ dimension-only RL
    & JA & 3 & \extraresult{71.15}{1.05} & \extraresult{84.71}{1.56} & \extraresult{51.94}{0.48} & \extraresult{45.72}{1.39} \\
LLM-only SFT $+$ gated RL (GRPO)
    & JA & 3 & \extraresult{70.34}{0.91} & \extraresult{83.73}{0.34} & \extraresult{51.39}{1.73} & \extraresult{49.36}{1.64} \\
LLM-only SFT $+$ gated RL; audio path updated during RL
    & JA & 3 & \extraresult{70.00}{1.19} & \extraresult{83.53}{1.76} & \extraresult{50.83}{1.44} & \extraresult{50.96}{3.51} \\
\addlinespace[1pt]
OPD initialization $+$ gated RL; audio path frozen
    & JA & 3 & \extraresult{73.91}{0.40} & \extraresult{85.88}{0.59} & \extraresult{56.94}{1.27} & \extraresult{52.87}{1.74} \\
OPD initialization $+$ gated RL; audio encoder updated
    & JA & 3 & \extraresult{72.76}{1.50} & \extraresult{85.10}{1.36} & \extraresult{55.28}{1.73} & \extraresult{53.99}{1.13} \\
\bottomrule
\end{tabular*}
\endgroup
\end{table}

\subsection{External Cross-Lingual Evaluation}
\label{app:crosslingual_evaluation}

We additionally evaluate on the public VOX-DUB benchmark~\citep{voxdub}, which
contains pairwise candidate comparisons from commercial cross-lingual speech
systems, with three human judgments per pair. Its five attributes map approximately to our rubric: voice,
emotion, naturalness, pronunciation, and sound quality correspond to Speaker,
Emotion, Timing, Pronunciation/Accent, and Audio Artifacts, respectively.  The
naturalness--Timing correspondence is the weakest, and VOX-DUB provides no
Overall verdict. We therefore treat the benchmark as an out-of-distribution
validity check rather than a second version of our primary evaluation.

For the English-to-Spanish subset, 21 utterances form 126 system
comparisons.  Because each commercial system occupies a fixed candidate slot
in the released data, we evaluate every comparison in both A/B orders and pool
the resulting 252 predictions.  Table~\ref{tab:voxdub_external} reports
three-class macro-F1.  Human lower and upper references score a rater against
the majority of the other raters or of all raters, respectively; with only
three annotations, these form a more honest bracket than a single human
ceiling.  Always-\textsc{Tie} and random baselines are important because the benchmark
is highly \textsc{Tie}-heavy, particularly for voice.

\begin{table*}[t]
    \centering
    \caption{External evaluation on English-to-Spanish VOX-DUB. Values are
    dimensional macro-F1 (\%); D3 (Naturalness) is an approximate
    match to our Timing dimension.}
    \label{tab:voxdub_external}
    \small
    \setlength{\tabcolsep}{6pt}
    \begin{tabular}{@{}lrrrrrr@{}}
        \toprule
        System & D1 Voice & D2 Emotion & D3 Naturalness$^*$ & D4 Pronun. & D5 Audio & Mean \\
        \midrule
        Human upper reference & 72.3 & 78.4 & 75.8 & 77.8 & 79.7 & 76.8 \\
        Human lower reference & 50.0 & 56.4 & 49.0 & 64.3 & 65.5 & 57.0 \\
        Always \textsc{Tie} & 29.5 & 11.4 & 22.6 & 25.9 & 19.9 & 21.8 \\
        Random & 25.8 & 31.1 & 31.9 & 29.6 & 32.5 & 30.2 \\
        \midrule
        Gemini 3.1 Pro & 26.1 & 45.1 & 35.9 & 45.0 & 48.2 & \textbf{40.0} \\
        EN--JA audio-adapted SFT & 19.8 & 38.9 & 36.2 & 51.8 & 22.5 & 33.8 \\
        English--Spanish SFT & 28.7 & 39.1 & 33.4 & \textbf{53.8} & 38.4 & 38.7 \\
        Joint-language SFT & 25.4 & \textbf{45.5} & 31.4 & 51.6 & \textbf{40.3} & 38.8 \\
        \bottomrule
    \end{tabular}
\end{table*}

The English--Spanish and joint-language judges reach mean macro-F1 of 38.7 and
38.8, close to Gemini-3.1-Pro's 40.0, and both exceed Gemini on pronunciation,
the dimension most directly targeted by our supervision.  No evaluated model
beats the Always-\textsc{Tie} baseline on voice, whose gold labels are 80\% \textsc{Tie} and have
low inter-rater reliability; the Voice dimension (D1) therefore supplies little discriminative signal
on this subset.  These results support transfer beyond our constructed test
sets while also showing why benchmark-specific label distributions and rubric
differences must remain visible.

\section{Human Evaluation of Diagnostic Rationales}
\label{app:rationale_evaluation}

\subsection{Protocol}

Because a diagnostic judge should support its verdicts with audibly verifiable
evidence localized to relevant words, phrases, or moments in the speech, we
treat evidence grounding and localization as central evaluation criteria.
We sample 244 unique rationale comparisons from the English--Japanese test set
of 290 comparisons and add 22 repeated assignments for quality control.
The two models in every comparison predict the same Overall verdict, preventing a
rater from preferring an explanation merely because it accompanies the more
accurate Overall verdict. The evaluation set includes 73 comparisons where both responses match
the human Overall label and all scored dimension-level labels, 122 where both match the
Overall label but at least one dimension-level label differs, and 49 where both
make the same incorrect Overall prediction.

Ten Japanese-speaking raters first listen to the English reference and both
Japanese candidates, then read two anonymous rationales in randomized order.
Model identities and automatic scores are hidden, and the instructions
explicitly separate rationale quality from the rater's preferred speech
candidate. Of the resulting 266 assignments, one is incomplete, leaving 265
complete judgments. The English wording shown in the annotation
interface is reproduced below; the deployed interface also included Japanese
translations.

\begin{figure}[!tbp]
\begin{promptbox}[before skip=0pt,after skip=0pt]{Instructions for Human Evaluation of Rationales}
\small
\begin{itemize}
    \setlength{\itemsep}{2pt}
    \setlength{\topsep}{4pt}
    \item Listen to the reference and both candidates before reading either rationale.
    \item Judge only the quality of the explanation, not which candidate you would have chosen.
    \item Do not prefer a rationale because its final verdict agrees with your own opinion.
    \item A rationale that states a difference you cannot hear is worse than one that stays silent.
\end{itemize}

\textbf{Comparison questions.}
For each question, choose \emph{Rationale X is better}, \emph{Rationale Y is
better}, \emph{Both are good / no meaningful difference}, or \emph{Both are
poor}.
\begin{enumerate}
    \setlength{\itemsep}{2pt}
    \setlength{\topsep}{4pt}
    \item Which rationale more clearly explains the important differences
    between the candidates, and gives more useful directions for improvement?
    \item Which rationale supports its claims with more accurate, specific,
    audibly verifiable evidence (cited words or phrases)?
    \item Which rationale's dimension-level observations more logically support
    its dimension-level judgments and its Overall verdict?
    \item Which rationale assigns observations to the correct evaluation
    dimensions more consistently, and avoids conflating different kinds of issues?
\end{enumerate}

\textbf{Overall rationale preference.}
Overall, which rationale is the better diagnostic explanation? Choose
\emph{Rationale X}, \emph{Rationale Y}, or \emph{Equally good}.

\textbf{Required comment.}
Briefly say why. One or two sentences.
\end{promptbox}
\end{figure}

\subsection{Pairwise Results}
\begin{table*}[b]
    \centering
    \caption{Human pairwise rationale evaluation. Each cell is
    \emph{first model better / second model better / no difference}; all paired
    models predict the same Overall verdict.}
    \label{tab:human_rationale_evaluation}
    \small
    \setlength{\tabcolsep}{4.5pt}
    \renewcommand{\arraystretch}{1.08}
    \begin{tabularx}{\textwidth}{@{}Xccccc@{}}
        \toprule
        Models & Overall & Communication & Grounding & Consistency & Dim. hygiene \\
        \midrule
        SFT vs. SFT+RL & 3/\textbf{11}/14 & 2/3/23 & 1/\textbf{11}/16 & 1/\textbf{6}/21 & 2/5/21 \\
        SFT vs. Gemini 3.1 Pro & \textbf{10}/7/9 & 3/6/17 & 6/6/14 & 1/1/24 & \textbf{9}/1/16 \\
        SFT vs. audio-adapted SFT & \textbf{10}/6/13 & 1/3/25 & \textbf{6}/3/20 & 2/1/26 & \textbf{5}/4/20 \\
        OPD (Qwen3 teacher) vs. OPD (SFT teacher)
        & \textbf{15}/5/5 & \textbf{5}/4/16 & \textbf{13}/2/10 & \textbf{5}/4/16 & \textbf{6}/5/14 \\
        \bottomrule
    \end{tabularx}
\end{table*}

Table~\ref{tab:human_rationale_evaluation} reports comparisons after mapping the
anonymous presentation order back to model identity. The four model pairings
shown account for 108 of the 265 complete judgments; the remaining 157 cover
auxiliary baseline and ablation pairings not used for the claims below.
``No difference'' pools
the two responses in which raters found both rationales comparably good or
comparably poor.  The SFT--RL comparison supports the main-text finding: most
judgments find the rationales similar, but perceived differences favor RL,
especially for audible-evidence grounding.  For evidence grounding, 11 of the
12 directional judgments favor SFT+RL over SFT (91.7\%; two-sided exact sign
test \(p=0.006\)); the remaining 16 judgments find no meaningful difference.
The Qwen3-teacher OPD rationales
receive an even stronger grounding preference over SFT-teacher OPD, despite
often accompanying incorrect dimension-level or Overall verdicts. This is why we
do not interpret textual detail or persuasiveness as evidence of diagnostic
correctness.

Agreement on repeated comparisons is 27\% for overall rationale preference and ranges
from 41\% to 64\% across the four criteria.  We therefore use this study to
identify large, coherent patterns and illustrative failure modes, not to claim
a complete ranking of explanation quality.  Figure~\ref{fig:rationale_example}
shows one abbreviated comparison.  We also tested a text-only GPT-5.4 grader,
but its localization score was only weakly predictive of human grounding
preferences (AUC \(=0.53\)), while its localized-cue count was no better than
chance (AUC \(=0.50\)).  Because a text-only grader cannot verify what is
audible, we do not report its model rankings or use them as evidence of
grounding; all grounding conclusions instead rely on listeners who inspected
the audio.

\end{document}